\documentclass{article} % For LaTeX2e
\usepackage{iclr2024_conference,times}

\usepackage{amsmath,amsfonts,bm}

\def\eqref#1{equation~\ref{#1}}
\def\1{\bm{1}}

\DeclareMathAlphabet{\mathsfit}{\encodingdefault}{\sfdefault}{m}{sl}
\SetMathAlphabet{\mathsfit}{bold}{\encodingdefault}{\sfdefault}{bx}{n}

\usepackage{url}

\usepackage[pagebackref,breaklinks,colorlinks]{hyperref}

\usepackage[utf8]{inputenc} % allow utf-8 input
\usepackage[T1]{fontenc}    % use 8-bit T1 fonts

\usepackage{booktabs}
\usepackage{graphicx}
\usepackage{bbm}
\usepackage{caption}
\usepackage{wrapfig}
\usepackage{makecell}

\usepackage{pifont}

\usepackage{cite}
\usepackage{amsmath,amssymb,amsfonts}
\usepackage{textcomp}
\usepackage{subcaption}
\usepackage{tabularx}
\usepackage{booktabs}       % professional-quality tables
\usepackage[normalem]{ulem}
\useunder{\uline}{\ul}{}
\usepackage{bm}
\usepackage{color}
\usepackage{xcolor}
\usepackage{algorithm}
\usepackage{algorithmicx}
\usepackage{algpseudocode}
\usepackage{multirow}
\usepackage{xspace}

\usepackage[utf8]{inputenc}
\usepackage[T1]{fontenc}
\usepackage[margin=1in]{geometry}
\usepackage{times}

\usepackage{algpseudocode}
\usepackage{enumitem}
\usepackage{lipsum}
\usepackage{hyperref}

\usepackage{booktabs}
\usepackage{colortbl}
\usepackage{xcolor}
\usepackage{multirow}

\usepackage{tcolorbox}
\tcbuselibrary{breakable}

\definecolor{lightgray}{gray}{0.9}
\definecolor{Gray}{gray}{0.9}

\usepackage{adjustbox}

\hypersetup{
    colorlinks=true,
    linkcolor=blue,
    filecolor=magenta,      
    urlcolor=cyan,
    citecolor=blue,
}

\newcommand{\symboletongyi}{\raisebox{0pt}{~\includegraphics[scale=0.05]{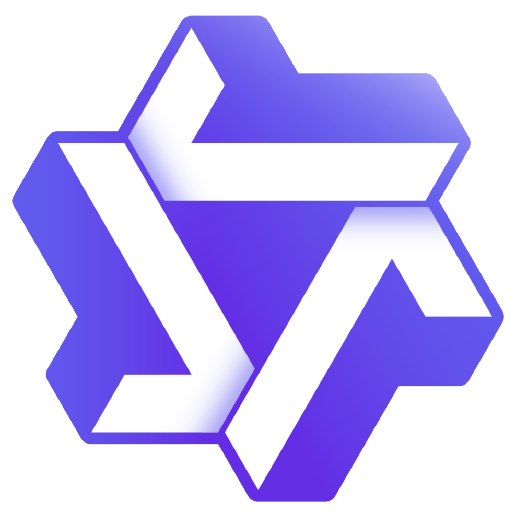}}~}

\usepackage{threeparttable}
\usepackage{newtxtext}
\usepackage[capitalize]{cleveref}
\crefname{section}{Sec.}{Secs.}
\Crefname{section}{Section}{Sections}
\Crefname{table}{Table}{Tables}
\crefname{table}{Tab.}{Tabs.}

\definecolor{myblue}{rgb}{0.2,0.2,0.6}
\definecolor{demphcolor}{RGB}{144, 144, 144}
\definecolor{mygray}{gray}{0.4}
\definecolor{lightgray}{rgb}{0.9, 0.9, 0.9}
\definecolor{deepgreen}{RGB}{0,100,0}

\newcommand{\cmark}{\color{mygray}\ding{51}}
\newcommand{\xmark}{\color{mygray}\ding{55}}

\hypersetup{%
  citecolor=teal
}
\hypersetup{linkcolor = black}
\newcommand{\modelname}{GUI-Owl-1.5\xspace}

\title{Mobile-Agent-v3.5: Multi-platform Fundamental GUI Agents}
\usepackage[symbol]{footmisc}
\iclrfinalarxiv % Uncomment this line to enable the non-anonymous mode
\begin{document}

\author{
}

\maketitle

\begin{center}
   \centering
   \vspace{-18mm}
   \textbf{Haiyang Xu\footnote{Core Contributors}\footnotemark[2]\qquad Xi Zhang\footnotemark[1] \qquad Haowei Liu\footnotemark[1] \qquad Junyang Wang\footnotemark[1] \qquad Zhaoqing Zhu\footnotemark[1] \qquad Shengjie Zhou \qquad  Xuhao Hu \qquad Feiyu Gao \qquad Junjie Cao \qquad Zihua Wang \qquad Zhiyuan Chen \qquad Jitong Liao \qquad Qi Zheng  \qquad Jiahui Zeng  \qquad Ze Xu \qquad Shuai Bai \qquad Junyang Lin \qquad Jingren Zhou \qquad Ming Yan \footnote{Corresponding author and project leader}} \\
    {Tongyi Lab\symboletongyi, Alibaba Group}\\
    {\tt\small \{shuofeng.xhy, ym11960\}@alibaba-inc.com} 
   
   \url{https://github.com/X-PLUG/MobileAgent}
\end{center}

\begin{abstract}
\vspace{-1mm}
The paper introduces GUI-Owl-1.5, the latest native GUI agent model that features instruct/thinking variants in multiple sizes (2B/4B/8B/32B/235B) and supports a range of platforms (desktop, mobile, browser, and more) to enable cloud-edge collaboration and real-time interaction. GUI-Owl-1.5 achieves state-of-the-art results on more than 20+ GUI benchmarks on open-source models: (1) on GUI automation tasks, it obtains 56.5 on OSWorld, 71.6 on AndroidWorld, and 48.4 on WebArena; (2) on grounding tasks, it obtains 80.3 on ScreenSpotPro; (3) on tool-calling tasks, it obtains 47.6 on OSWorld-MCP, and 46.8 on MobileWorld; (4) on memory and knowledge tasks, it obtains 75.5 on GUI-Knowledge Bench. GUI-Owl-1.5 incorporates several key innovations: (1) \textbf{Hybird Data Flywheel}: we construct the data pipeline for UI understanding and trajectory generation based on a combination of simulated environments and cloud-based sandbox environments, in order to improve the efficiency and quality of data collection. (2) \textbf{Unified Enhancement of Agent Capabilities}: we use a unified thought-synthesis pipeline to enhance the model's reasoning capabilities, while placing particular emphasis on improving key agent abilities, including Tool/MCP use, memory and multi-agent adaptation; (3) \textbf{Multi-platform Environment RL Scaling}: We propose a new environment RL algorithm, MRPO, to address the challenges of multi-platform conflicts and the low training efficiency of long-horizon tasks. The GUI-Owl-1.5 models are open-sourced, and an online cloud-sandbox demo is available at https://github.com/X-PLUG/MobileAgent.

\end{abstract}
\begin{figure}[H]
    \centering
    \vspace{-0.5cm}
    \includegraphics[width=0.99\textwidth]{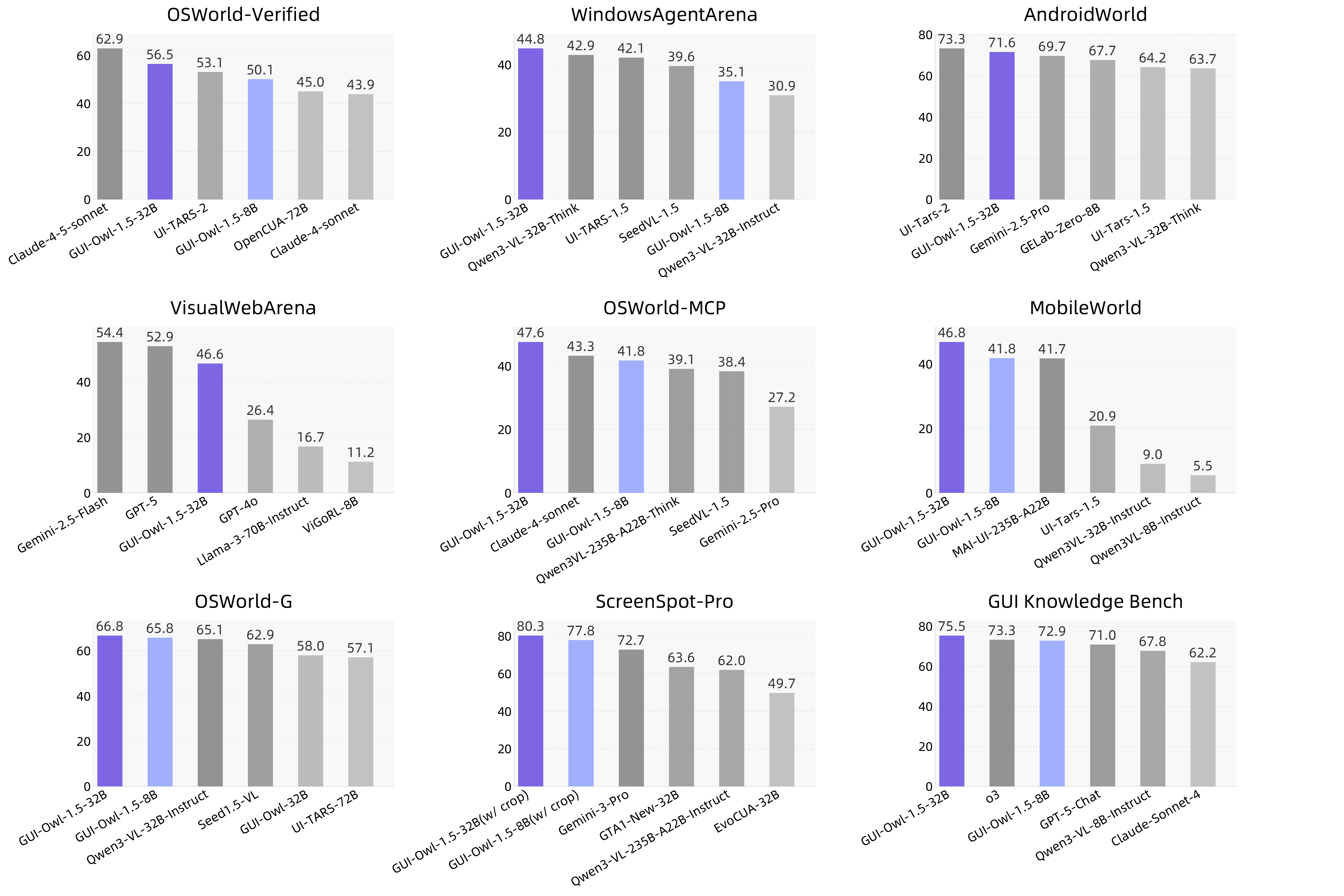}
    \vspace{-0.2cm}
    \caption{Performance overview on mainstream GUI task automation, grounding and knowledge benchmarks.}
\label{fig:overall_results}
% \vspace{-0.8cm}
\end{figure}

% \begin{figure}[h]
%     \centering
%     % \includegraphics[width=\textwidth]{images/score.pdf}
%     \includegraphics[width=\textwidth]{images/score_new.pdf}
%     \caption{Performance overview on mainstream GUI-automation benchmarks.}
%     \label{fig:score}
% \end{figure}

\begin{figure}[t]
    \centering
    \includegraphics[width=0.9\textwidth]{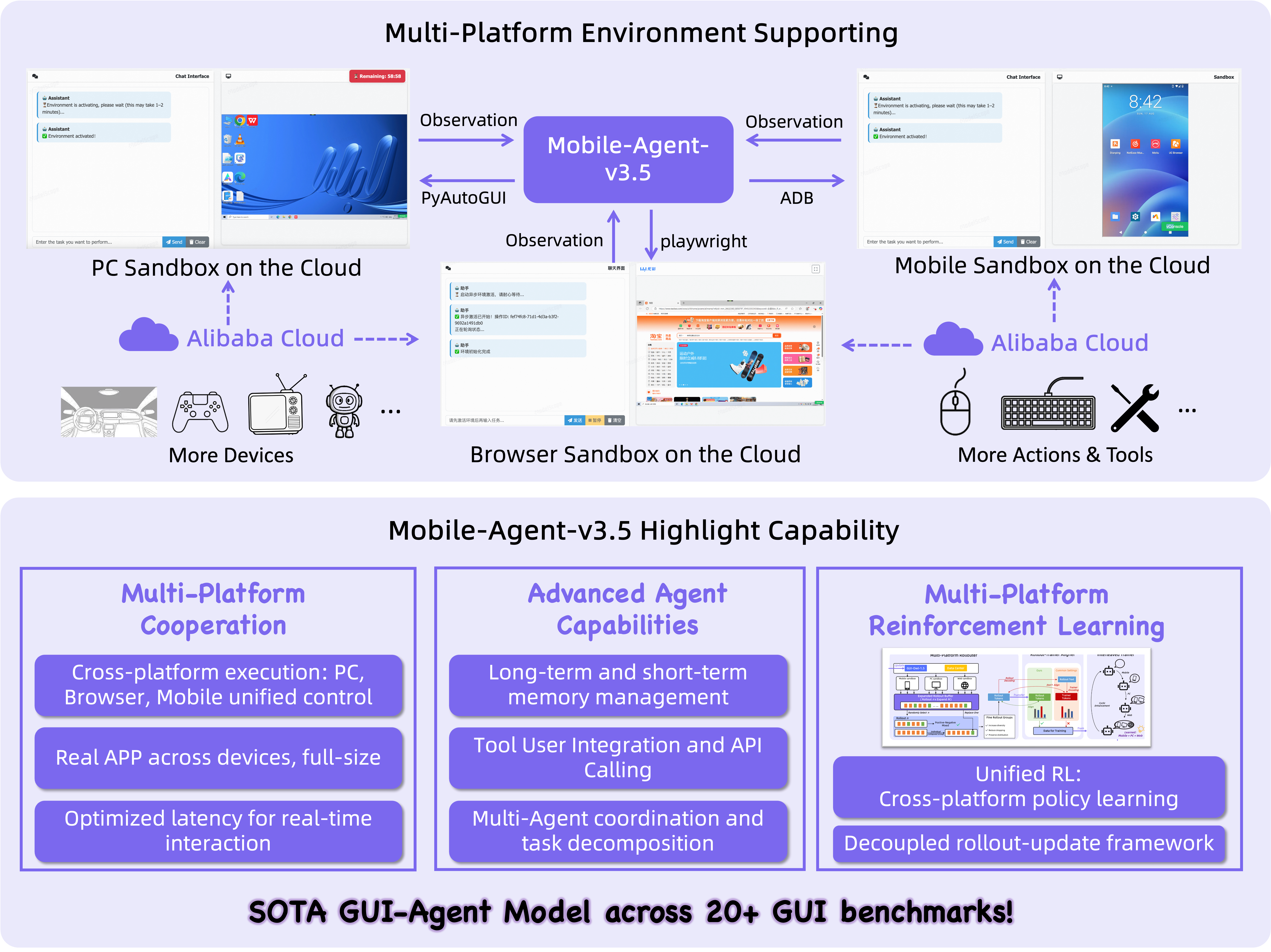}
    % \vspace{-0.1cm}
    \caption{Overview of our Mobile-Agent-v3.5. We illustrate our multi-platform environment supporting and our highlight
capability.}
    \label{fig1}
    % \vspace{-0.25cm}
\end{figure}

\section{Introduction}
With the rapid development of Vision–language models (VLMs)~\citep{Qwen3-VL, anthropic2025claude4, openai2025gpt5, gemini3pro},multimodal agents ~\citep{wang2024mobile, qin2025ui, ye2025mobile, liu2024autoglm, zhou2025mai, opencua}have achieved substantive progress, especially Graphical user interface (GUI) agents. GUI Agents are mainly designed to perform automated operations across multiple devices, such as desktops, mobiles, browsers, and so on. Recently, native agent models~\citep{ye2025mobile, qin2025ui, ui-tars-2-seed} based on end-to-end learning have demonstrated great potential, rather than only building agent frameworks on top of closed-source models~\citep{wang2024mobile2, wang2025mobile, Agent-S2, zhang2025appagent}.

However,the development of robust and practically usable GUI agents still faces several challenges. (1) The efficiency of real-world data collection: Collecting large-scale trajectories is costly to hamper the scalability of GUI datasets, as it requires complex agentic workflows, manual annotation, and engineering-level handling of anomalous scenarios; (2) The adaptation to multiple platforms: The native agent model needs to perform automated tasks reliably across a wide range of devices, including desktops, mobiles, browsers, and in-vehicle systems. It should also support complex agentic real-time interactions, such as edge–cloud collaboration and coordination across multiple devices; (3) The comprehensive agentic capabilities: The General GUI Agent should be capable of completing tasks efficiently, not limited to GUI-only operations. It should also support tool/Model Context
Protocol (MCP) invocation, short-term and long-term memory, multi-agent adaptation, and human–agent interaction.

To address these challenges, we propose GUI-Owl1.5, our latest native GUI agent model for multi-platform GUI automation across desktops, mobiles, browsers, and more. Built on Qwen3-VL and powered by a scalable data pipeline and a multi-stage training paradigm, GUI-Owl1.5 comprises a family of foundation GUI models covering a full range of sizes, including instruct/thinking variants at 2B, 4B, 8B, 32B, and 235B-A22B. Smaller instruct models, which do not produce thoughts, enable faster inference and can be deployed on edge devices to support high-frequency, real-time interactions while addressing security and privacy concerns. Larger thinking models, with stronger capabilities in task planning and reflection, are better suited for complex tasks and can collaborate with edge-deployed instruct models in a multi-agent setup to enable edge–cloud collaboration and multi-platform coordination. The key technical points are highlighted next.

\textbf{Hybird Data Flywheel:}
We develop the data pipeline for UI understanding and trajectory generation by synergistically integrating simulated environments with cloud-based platform environments, thereby enhancing both the efficiency and quality of data collection. For Grounding: a comprehensive grounding data augmentation pipeline that encompasses both hard grounding data generation—including challenging app GUI synthesis and multi-window high-resolution scenarios—and scalable high-quality data extension through trajectory mining, tutorial knowledge extraction, and infeasible query generation. For trajectory, we build a self-evolving trajectory synthesis workflow based on a directed acyclic graph (DAG) \citep{ye2025mobile}. Meanwhile, we synthesize virtual environments via Vibe Coding to create high-frequency, complex atomic operations and apps featuring challenging cases such as pop-ups and CAPTCHA-style verifications. In addition, for some challenging apps and scenarios, we incorporate a small amount of manual annotation to better align synthetic environments with real-world ones.

\textbf{Unified Enhancement of Agent Capabilities:} Beyond basic GUI perception and action execution, a practical GUI agent must possess a range of higher-order skills. We introduce three complementary strategies to comprehensively enhance the native model's agent capabilities. First, we inject GUI knowledge through large-scale QA data crawled from software documentation and forums, and train the model with world modeling supervision to anticipate interface state transitions before acting. Second, we design a unified chain-of-thought (CoT) synthesis pipeline that augments all trajectory data with step-wise observation, reflection, memory management, and tool invocation reasoning, enabling superior long-horizon planning and in-context information retention. Third, we incorporate multi-agent collaboration data collected via the Mobile-Agent-v3.5 framework, allowing the model to function not only as a standalone end-to-end agent but also as specialized roles (\textit{e.g.}, planner, executor, verifier) within structured multi-agent systems.
% To build a fundamental GUI agent, we comprehensively enhance the native GUI model’s capabilities, including planning and reflection, MCP/tool calling, short-term memory, compatibility with multi-agent frameworks, and world-knowledge understanding.

\textbf{Multi-platform Environment RL Scaling:}
To enable stable reinforcement learning training across multi-platform environments, we propose MRPO (Multi-platform Reinforcement Policy Optimization), a large-scale RL framework that addresses four critical challenges in GUI agent training. First, we unify learning across mobile, desktop, and web environments under a single device-conditioned policy. Second, we introduce an online rollout buffer that mitigates GRPO training instability when grouped rollouts collapse to identical outcomes by oversampling trajectories and strategically selecting diverse groups while maintaining on-policy guarantees. Third, we ensure consistency between environment-side inference and training-side optimization through token-ID transport, preventing tokenization mismatches. Finally, we adopt alternating multi-platform optimization to reduce gradient interference, training on single device types cyclically rather than mixing trajectories. This approach enables stable, unified policy learning while preserving cross-device generalization for long-horizon GUI control tasks.

% We evaluate GUI-Owl1.5 across a wide range of benchmarks that comprehensively measure native agent capability
% on GUI automation including GUI navigation, grounding, tool calling, memory, and knowledge tasks.
We evaluate GUI-Owl-1.5 on a series of benchmarks spanning GUI task automation, grounding, tool invocation, memory and knowledge.
Experimental results demonstrate that GUI-Owl-1.5 exhibits strong GUI understanding, grounding and execution capabilities, achieving state-of-the-art performance among open-source models across more than 20 GUI benchmarks. Specifically, it attains task success rates of 56.5\%, 71.6\% and 46.6\% on OSWorld-Verified, AndroidWorld and VisualWebArena respectively, which outperforms models such as UI-TARS-2, Claude-4, and Gemini -2.5-Pro.
On OSWorld-MCP, which evaluates the integration of GUI operations and tool invocation, it achieves a task success rate of 47.6\%. 
% On the ScreenSpot-Pro grounding benchmark, it achieves an accuracy of 80.3\% combined with crop. 
On the ScreenSpot-Pro grounding benchmark, it achieves a state-of-the-art accuracy of 80.3\% with crop-based refinement, and notably surpasses the large-scale Gemini-3-Pro even in its base configuration (72.9\%) without crop tool.
On MemGUI-Bench and GUI Knowledge Bench, our model also surpasses previous open-source models.

\section{Mobile-Agent-v3.5}

GUI-Owl-1.5 is a multimodal model for GUI operations, building on the previous GUI-Owl~\citep{ye2025mobile}. Compared to its predecessor, it offers three main improvements: (1) a broader action space; (2) improved context retention; (3)enhanced design in synthetic data generation, cross-platform adaptation, and agent capabilities.
% 

% GUI-Owl-1.5 is an end-to-end multimodal GUI foundation model that extends the GUI-Owl~\ref{} with a broader action space, improved context retention, and enriched designs emphasizing synthetic environments, multi-end adaptation, and comprehensive agentic capabilities.
Building on Qwen3-VL and trained with extensive post-training datasets, GUI-Owl-1.5 maintains the core functions of the original model—perceiving, planning, decision-making, and locating elements in GUI scenarios—while further optimizing them for different real-world cases. The model can autonomously interact with mobile, desktop, and browser interfaces across multiple turns, and can also work collaboratively in multi-agent systems.

\subsection{Formulation}
We formulate the GUI agent task as a multi-turn interactive decision-making problem, where the agent continuously perceives the environment, executes actions, and adapts its strategy based on real-time feedback.

\textbf{Input Space.} At each interaction step $t$, the agent receives:
\begin{itemize}
\item Visual observation $\mathcal{I}_t \in \mathbb{R}^{H \times W \times 3}$: a screenshot capturing the current GUI state.
\item User instruction $\mathcal{L}_t$: a natural language command expressing the user's intent.
\end{itemize}

\textbf{Output Space.} Given the input, the agent generates:
\begin{itemize}
\item Action conclusion $\mathcal{C}_t$: a natural language explanation summarizing the planned action.
\item Tool call $\mathcal{A}_t$: a structured function call that executes the action.
\end{itemize}

\begin{figure}[!t]
    \centering
    \includegraphics[width=0.99\textwidth]{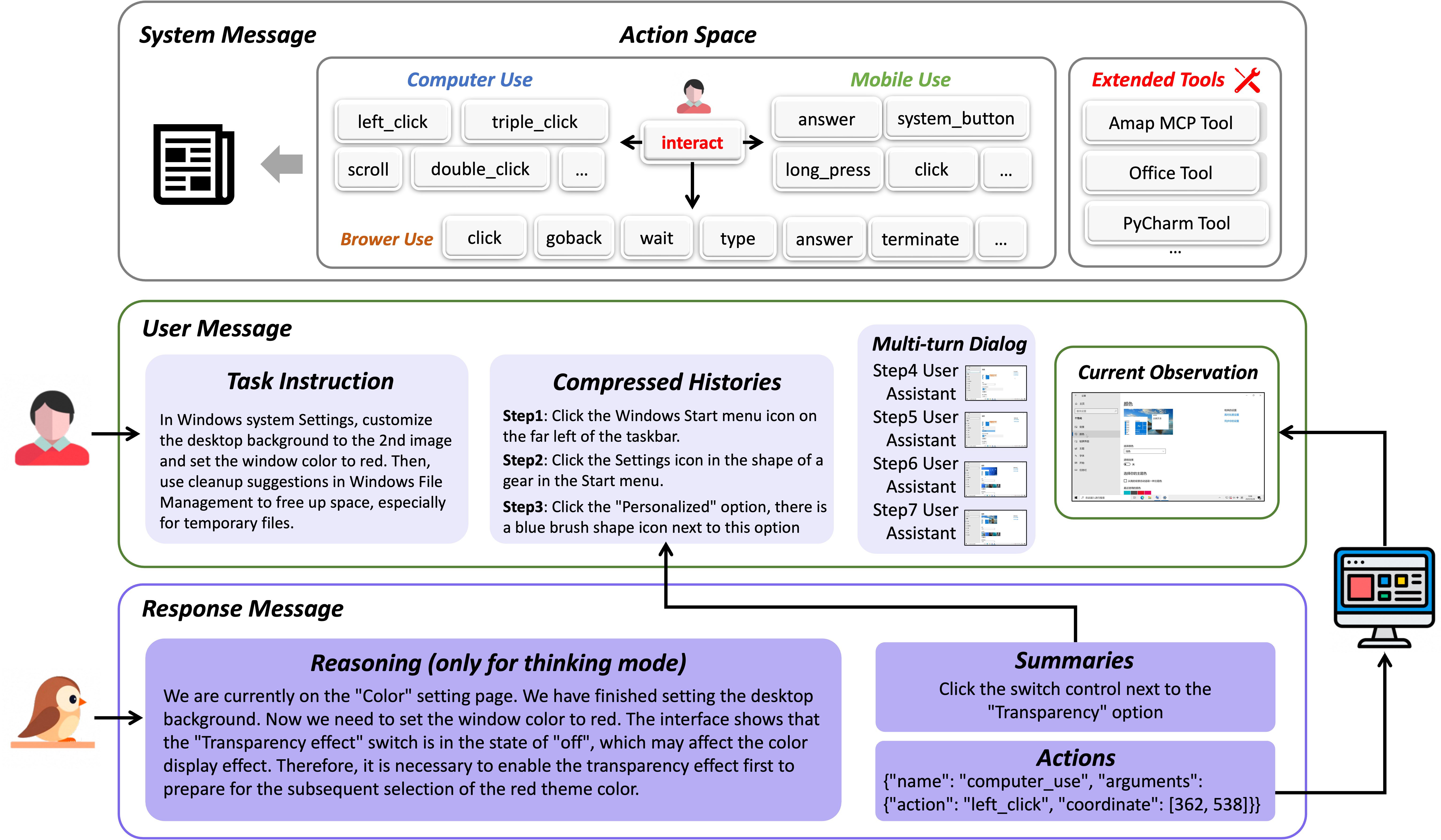}
    % \vspace{-0.1cm}
    \caption{Illustration of the interaction flow of GUI-Owl-1.5. The system message defines the available action space,
the user message contains the task instruction, compressed histories, and current observation, while the response
message includes the agent’s reasoning, action summaries, and the final action output.}
    \label{fig2}
    % \vspace{-0.25cm}
\end{figure}

The nature of GUI agent tasks requires closed-loop interaction with the environment. Specifically, after executing $\mathcal{A}_t$, the environment transitions to a new state, providing updated visual feedback $\mathcal{I}_{t+1}$ for the next turn. This iterative process continues until the task is completed or terminated.
Notably, compared to the previous GUI-Owl, we significantly expand the action space to support \textbf{external tool calls} and \textbf{API invocations} in addition to primitive GUI operations (\textit{e.g.}, click, type, scroll). This extension enables the agent to orchestrate complex workflows across heterogeneous systems, such as querying databases through APIs, invoking specialized computational tools, and integrating with third-party services. 

% The complete action space is detailed in Table~X.

\textbf{Context Management.} To handle long-horizon tasks while maintaining computational efficiency, we adopt a sliding window mechanism with hierarchical context compression. The context at step $t$ is organized as:

\begin{itemize}
\item Recent context (full retention): The most recent $N$ complete dialogue turns, including all modalities: $\{(\mathcal{I}_{t-N}, \mathcal{L}_{t-N}, \mathcal{C}_{t-N}, \mathcal{A}_{t-N}), \ldots, (\mathcal{I}_{t-1}, \mathcal{L}_{t-1}, \mathcal{C}_{t-1}, \mathcal{A}_{t-1})\}$

\item Historical context (compressed summary): Earlier interactions beyond the $N$-turn window are condensed into a textual summary $\mathcal{S}_{1:t-N-1}$, formed by concatenating action conclusions: $\mathcal{S}_{1:t-N-1} = \text{concat}(\mathcal{C}_1, \mathcal{C}_2, \ldots, \mathcal{C}_{t-N-1})$
\end{itemize}

This hierarchical design preserves fine-grained multi-modal information for immediate decision-making while maintaining high-level awareness of long-term task progression, effectively balancing context richness with memory efficiency.

% \subsubsection{Foundational Agent Capabilities}
% ----- refering -----

% Mobile-Agent-v3.5 introduces updated instruction templates emphasizing:

% Synthetic Environment Support – The model can operate within simulated GUI scenarios. Synthetic feedback loops enable pre-deployment validation, simulation-based training, and safe testing without impacting live systems.

% Multi-end Adaptation – Supports seamless task execution across Mobile, PC, and Web platforms. MCP Tool enables layered workflows where steps on different endpoints are combined into a unified execution plan.

% Comprehensive Agentic Capability – Maintains independent operation while also functioning cooperatively in multi-agent setups. Enhanced context retention plus MCP Tool integration allow the model to handle complex, distributed workflows requiring negotiation, coordination, and information synthesis.

\subsection{Data Preparation}
To support training of GUI-Owl-1.5 across heterogeneous platforms and task families, we develop a unified data preparation pipeline that targets both actionable interaction supervision and fine-grained visual grounding. Specifically, we curate (i) trajectory data that captures long-horizon decision-making and tool-augmented execution in realistic GUI environments, and (ii) grounding data that aligns natural-language intents with on-screen elements.

\subsubsection{Grounding}
\begin{figure}[!t]
    \centering
    \includegraphics[width=0.99\textwidth]{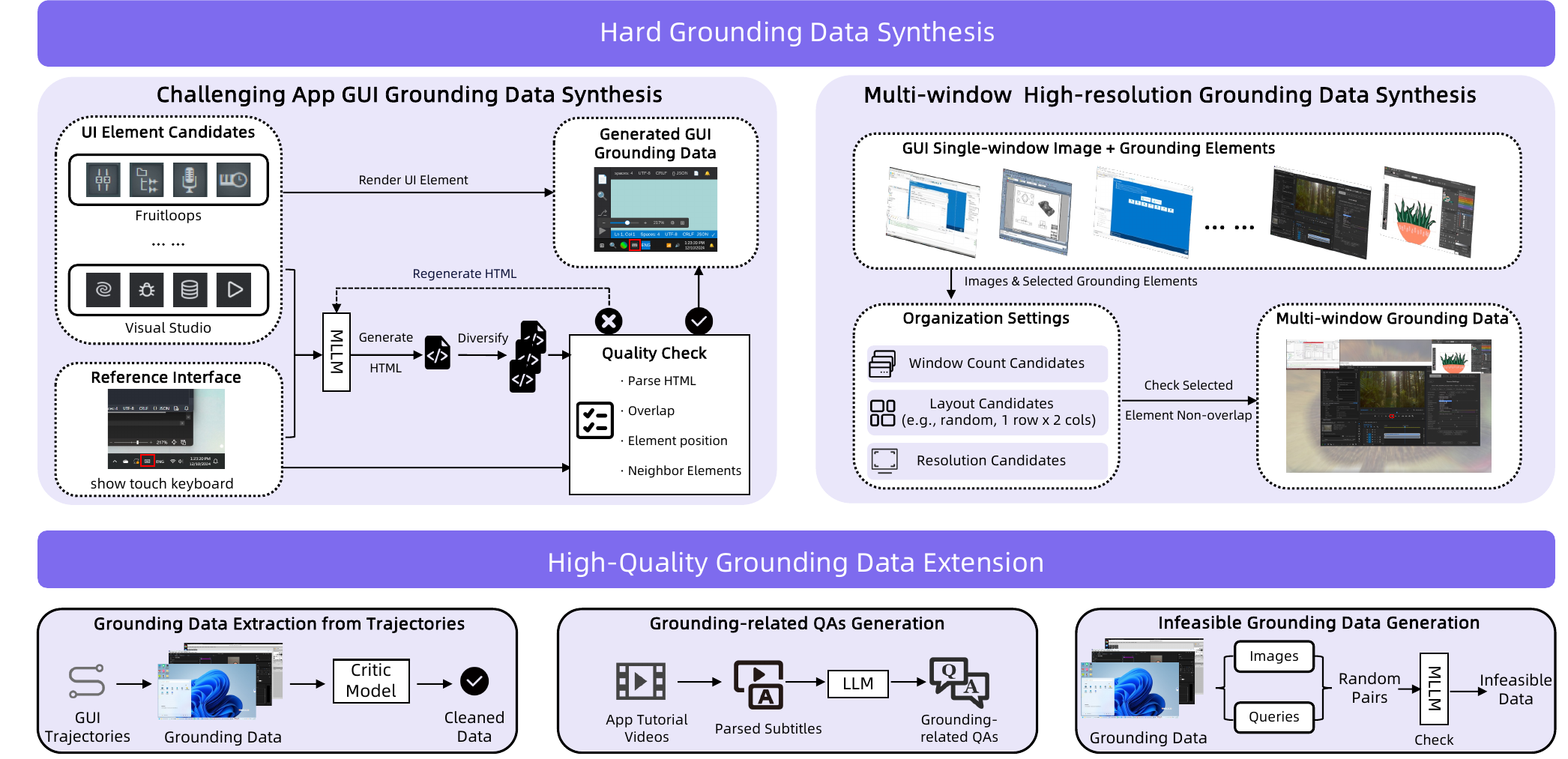}
    \vspace{-0.1cm}
    \caption{Overview of our high-quality grounding data construction pipeline.}
    \label{fig:grounding_pipeline}
    % \vspace{-0.25cm}
\end{figure}

% 现有的grounding数据相对简单且单一，稀缺高质量、高难度的grounding数据以及低成本data scaling的数据方案。为了解决该问题，如图xx所示，我们从两个方面来有效扩充grounding数据，以高效提升GUI的grounding能力
Existing grounding datasets exhibit limited complexity and diversity, creating a critical shortage of high-quality, challenging grounding data alongside scalable data augmentation solutions. As illustrated in Figure~\ref{fig:grounding_pipeline}, we address these limitations through a comprehensive data augmentation framework that enhances GUI grounding capabilities via two complementary strategies.

For \textbf{\textit{Hard Grounding Data Generation}}, which targets complex scenarios requiring specialized domain knowledge and high annotation costs, we develop two synthesis approaches:
\begin{itemize}
\item Challenging App GUI Grounding Data Synthesis: We leverage annotated UI elements and reference interfaces to generate diverse, high-quality professional application screenshots using MLLMs. This process incorporates iterative quality assessment and refinement mechanisms, where generated interfaces undergo validation checks and corrective regeneration to ensure data fidelity and domain accuracy. 

\item Multi-window High-resolution Grounding Data Synthesis: Utilizing existing single-window datasets combined with candidate organization pools (encompassing window count variations, layout configurations, and resolution options), we generate complex multi-window scenarios while ensuring target UI elements remain unoccluded through spatial constraint validation.
% 对于Hard Grounding数据（e.g., 专业软件、多窗口数据），仅靠human标注是非常难的，需要专业知识以及高成本。我们探索2种方式方式：
% (1) 探索专业app合成环境的高质量数据合成策略：利用已标注的UI element+参考界面，利用MLLM来生成大量的高质量多样化的专业app界面以及grounding数据，并进行质量check以及重新生成矫正来获取质量更高的数据
% (2) 多窗口高分辨率grounding数据合成：利用现在的单窗口数据以及候选的组织池（包括窗口数量、排布候选、分辨率候选），保证指定element不被遮挡的情况下生成高质量多窗口数据
\end{itemize}

For \textbf{\textit{High-Quality Grounding Data Extension}}, which aims to achieve cost-effective and scalable data augmentation, we implement three synergistic enhancement pathways:
\begin{itemize}
\item Trajectory-based grounding extraction: We mine grounding annotations from existing PC and mobile simulation environment trajectories, employing critic models to filter and validate data quality, ensuring only high-fidelity grounding pairs are retained.
\item Tutorial-based knowledge mining: Application tutorials are parsed to extract grounding-related question-answer knowledge by analyzing embedded subtitles and identifying spatial-semantic relationships, ultimately generating comprehensive grounding-oriented QA pairs that capture real-world usage patterns.
\item Infeasible query generation: To address the critical gap in handling infeasible grounding queries within existing datasets, we generate large-scale negative samples through strategic random pairing of queries and interface elements, followed by multi-model consensus filtering to identify and validate truly infeasible grounding instances.
% 另外，为了更低成本且更有效地扩充Grounding数据，我们考虑以下3种路径联合强化grounding信息：
% (1) 从已有的模拟pc、mobile环境中获取的轨迹数据中获取grounding数据，并通过critic model进行清洗
% (2) app教程里获取大量grounding相关的QA知识，我们通过解析里面的字幕并挖掘里面的grounding相关的grounding知识，最后获取最终的grounding-related QAs
% (3) 不可行数据生成：为了弥补现有的grounding数据对于不可行query的判断，利用random pair再用多模型筛选不可行样本来生成大量不可行grounding数据
\end{itemize}

\begin{figure}[!t]
    \centering
    \includegraphics[width=0.99\textwidth]{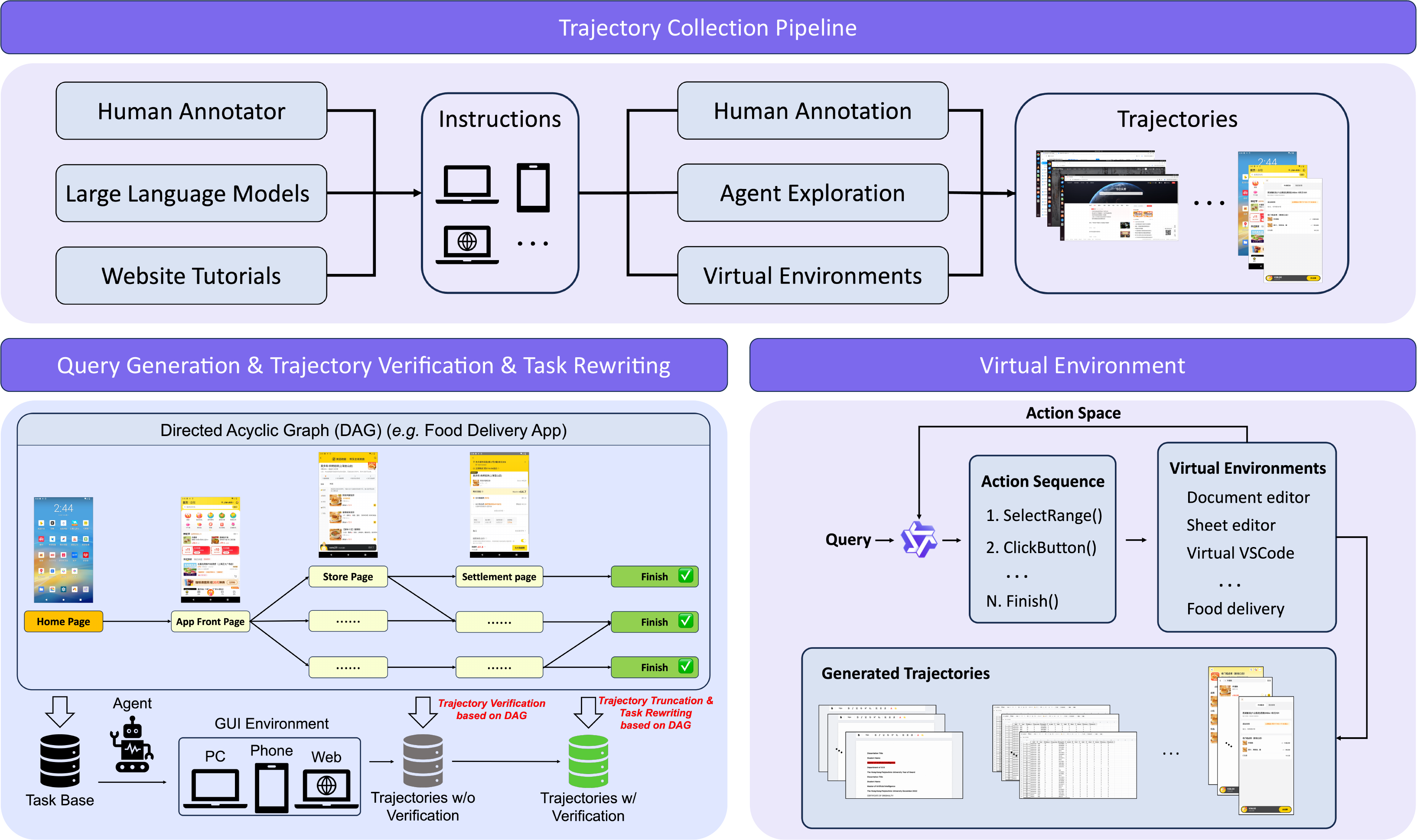}
    \vspace{-0.1cm}
    \caption{Overview of our trajectory collection pipeline.}
    \label{fig:data_pipeline}
    % \vspace{-0.25cm}
\end{figure}
\subsubsection{Trajectory Data Collection}
We build a hybrid trajectory corpus that scales to diverse applications and devices while maintaining high supervision fidelity. The pipeline consists of (i) DAG-based task synthesis to cover frequent workflows, (ii) automated rollouts on real devices with DAG-based validation, (iii) human demonstrations for tasks that remain unsolved by automation, (iv) trajectory production via virtual environments for basic actions (\textit{e.g.}, scroll, drag) and high-frequency challenging scenarios.
% simulated-environment generation for concentrated hard scenarios, and (v) a unified chain-of-thought (CoT) synthesis procedure that augments all trajectories with thought and conclusion signals.

\paragraph{Task production via human-authored DAGs.}
For each application domain, annotators construct a directed acyclic graph (DAG)
\[
G=(V,E),\quad V=\{v_i\}_{i=1}^{|V|},\quad E\subseteq V\times V,
\]
where each node \(v_i\) denotes an atomic subtask and each edge \((v_i,v_j)\in E\) denotes a feasible transition under typical UI state evolution. Let \(S\subseteq V\) and \(T\subseteq V\) be the sets of valid start and terminal nodes separately, we synthesize a task by sampling a path from \(S\) to \(T\):
\[
p=(v_1,\dots,v_K),\quad v_1\in S,\ v_K\in T,\ (v_k,v_{k+1})\in E,
\]
which represents a realistic action sequence with multiple steps. Each node \(v_k\) is associated with a sub-instruction template \(d(v_k)\) that optionally has slots for diverse entities. The final task instruction is composed by concatenating and rewriting the ordered sub-instructions:
\[
\mathcal{I}(p)=\operatorname{Compose}\big(d(v_1),d(v_2),\dots,d(v_K)\big).
\]
By sampling diverse paths and instantiating templates, the DAG provides controllable coverage of high-frequency operation patterns in common apps, minimizing the impact of the LLM hallucination.

\paragraph{Automated trajectory generation with checkpointing, truncation, and task repair.}
Given \(\mathcal{I}(p)\), an agent interacts with a real device environment \(\mathcal{E}\) to produce a trajectory
\[
\tau=\{(o_t,a_t)\}_{t=1}^{T},
\]
where \(o_t\) is the observation (e.g., screenshot, UI structure, device metadata) and \(a_t\) is the executed action (touch/keyboard/tool call). To assess partial completion along the subtask path \(p\), we define a checkpoint predicate for each node \(v_k\):
\[
\phi_{k}:\mathcal{O}\rightarrow\{0,1\},\quad \phi_k(o_t)=1\ \text{iff subtask }v_k\text{ is satisfied at }o_t.
\]
We compute whether subtask \(k\) is achieved somewhere in the rollout by
\[
c_k(\tau)=\max_{t\in[1,T]}\phi_k(o_t).
\]
The longest completed prefix length is
\[
m(\tau)=\max\left\{m\in\{0,\dots,K\}\ :\ \forall k\le m,\ c_k(\tau)=1\right\}.
\]
If \(m(\tau)=K\), we accept \(\tau\) as a correct trajectory. Otherwise, we truncate the rollout to the last verified checkpoint of the completed prefix:
\[
t^\star=\max\{t:\phi_{m(\tau)}(o_t)=1\},\quad \tau'=\{(o_t,a_t)\}_{t=1}^{t^\star},
\]
and repair the original task by removing the completed subtasks:
\[
p_{\text{rem}}=(v_{m(\tau)+1},\dots,v_K),\quad \mathcal{I}_{\text{rem}}=\mathcal{I}(p_{\text{rem}}).
\]
We then store \((\mathcal{I}_{\text{rem}},\tau')\) as a partially-correct instance, which provides clean supervision for the successfully executed segment while avoiding noisy labels beyond the last correct subtask.

\paragraph{Human annotation on real devices.}
For difficult tasks that remain unsolved after repeated automated attempts, we collect expert demonstrations via a cloud annotation platform. Annotators directly operate the same real device environments and record gold trajectories \(\tau^{\text{human}}\) aligned with the task instruction, ensuring high-quality supervision for hard cases.

% \paragraph{Virtual-environment trajectory production for concentrated hard scenarios.}
\paragraph{Virtual environment-based trajectory production.}
Relying solely on agent exploration in real-world environments for trajectory generation presents two notable limitations: (i) Real-world applications and software often incorporate CAPTCHA verification, anti-bot mechanisms, and other protective measures that can interrupt or terminate the agent's exploration process. (ii) Real-world environments cannot provide accurate feedback, which results in low efficiency of trajectory generation via agent exploration, and often yields trajectories that contain erroneous or redundant steps.
% In certain high-frequency scenarios, the inability of real-world environments to provide accurate feedback results in low efficiency of trajectory generation through agent exploration, often yielding trajectories that contain erroneous or redundant steps.

To address these challenges, we develop a suite of web-rendering-based virtual environments targeting fine-grained primitive actions (\textit{e.g.}, scroll, drag) and high-frequency difficult scenarios (\textit{e.g.}, document and spreadsheet editing, popular applications). These virtual environments serve two primary purposes: (i) providing precise sub-task-level feedback to guide agent exploration, and (ii) enabling automated and scalable trajectory generation through the integration of LLM-based instruction decomposition.

% When hard failures concentrate in a small number of recurring UI configurations (e.g., specific app versions, permission dialogs, locale-dependent layouts), real-device collection becomes inefficient. We therefore develop a simulator \(\tilde{\mathcal{E}}\) that supports (i) agent rollouts with a subtask-sequence critic for scalable filtering, and (ii) script/RPA execution for high-throughput generation of guaranteed-correct trajectories.

\emph{Agent rollout + critic.} Given a sampled scenario \(\omega\) and a DAG path \(p=(v_1,\dots,v_K)\), the agent produces a simulated trajectory \(\tilde{\tau}=\{(\tilde{o}_t,\tilde{a}_t)\}_{t=1}^{\tilde{T}}\). The simulator exposes subtask-completion predicates \(\tilde{\phi}_k(\tilde{s}_t)\in\{0,1\}\), enabling an exact prefix-progress score:
\[
\tilde{c}_k(\tilde{\tau})=\max_{t}\tilde{\phi}_k(\tilde{s}_t),\qquad
\tilde{m}(\tilde{\tau})=\max\left\{m:\forall k\le m,\ \tilde{c}_k(\tilde{\tau})=1\right\}.
\]
We accept \(\tilde{\tau}\) if \(\tilde{m}(\tilde{\tau})=K\); otherwise we truncate to the last verified checkpoint and keep a clean partially-correct prefix for training:
\[
\tilde{t}^\star=\max\{t:\tilde{\phi}_{\tilde{m}(\tilde{\tau})}(\tilde{s}_t)=1\},\qquad
\tilde{\tau}'=\tilde{\tau}_{1:\tilde{t}^\star}.
\]

\emph{Scalable Automated Trajectory Generation.} Since our virtual environments are built upon web rendering, they inherently support the automated execution of atomic operations. For a given virtual environment, such as a virtual word document editor, we leverage an LLM in conjunction with the document content to decompose a user instruction into a sequence of atomic operations that the virtual environment can execute. These atomic operations are then fed into the virtual environment to produce corresponding precise operation trajectories.
% \emph{Script/RPA trajectories.} 
Also, for many concentrated scenarios, the canonical correct operation is known and can be standardized. We encode it as a script/RPA policy \(\rho\) and directly execute:
\[
\tilde{\tau}^{\text{rpa}}=\operatorname{Rollout}(\tilde{\mathcal{E}},\rho,\omega,p),
\qquad \tilde{m}(\tilde{\tau}^{\text{rpa}})=K,
\]
which yields high-quality successful trajectories at low cost. 
% In practice, the above two modes are complementary: scripts provide reliable successes, while agent rollouts provide diverse behaviors that are efficiently filtered by the subtask critic.

\subsection{Agent Capability Enhancement}
\begin{figure}[!t]
    \centering
    \includegraphics[width=0.98\textwidth]{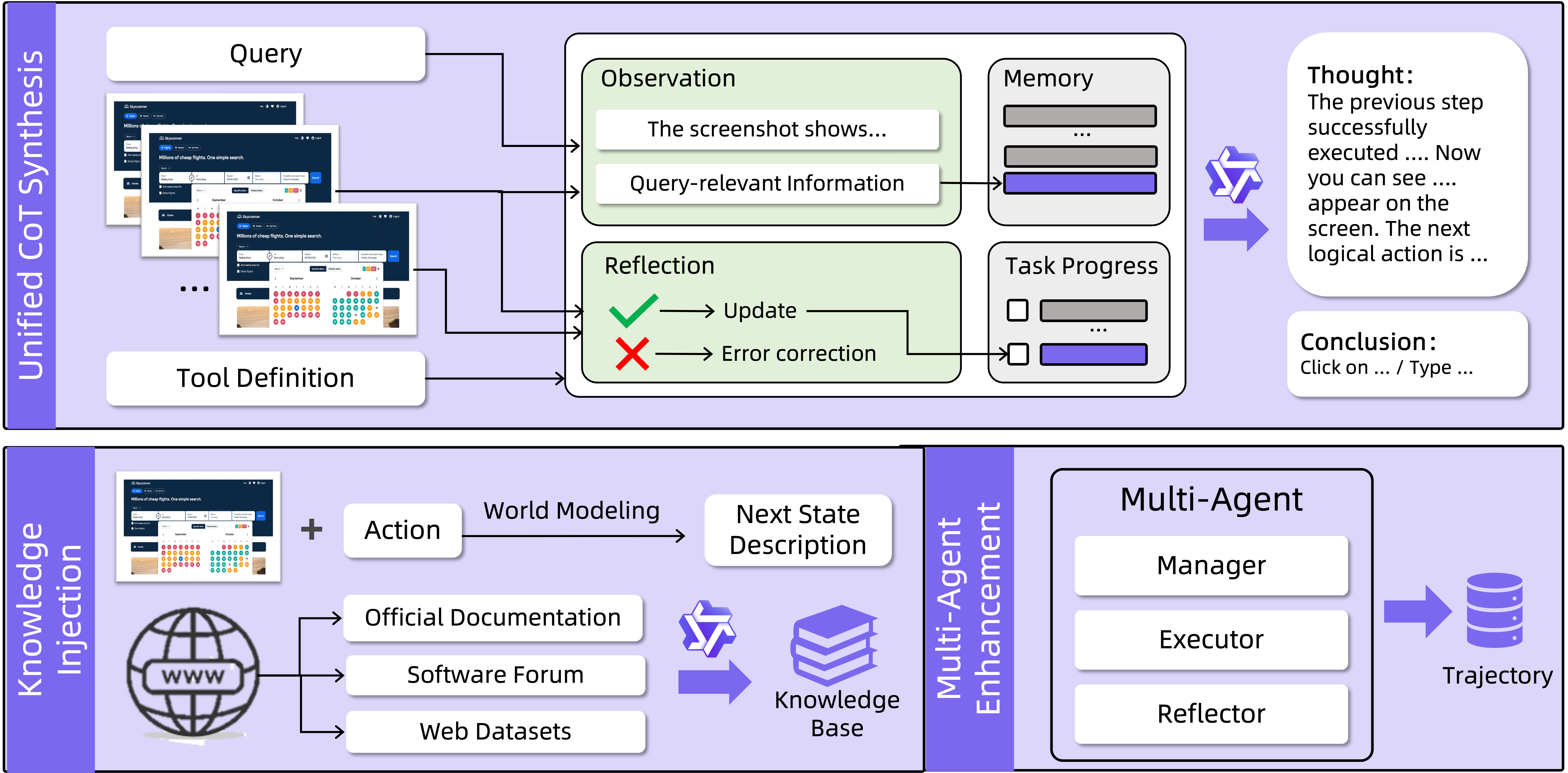}
    \vspace{-0.1cm}
    \caption{Illustration of our agent capability enhancement pipeline.}
    \label{fig:capability_enhancement}
    % \vspace{-0.25cm}
\end{figure}

Beyond grounding and GUI understanding, a capable GUI agent must plan over long horizons, reason about action consequences, memorize key information, and invoke external tools. We introduce three complementary strategies to enhance these capabilities (Figure~\ref{fig:capability_enhancement}): (i) {GUI Knowledge Injection}, which enriches the model's knowledge through QA data and world modeling; (ii) {Unified CoT Synthesis}, which augments trajectory data with step-wise reasoning, reflection, and memory; and (iii) {Multi-Agent Collaboration}, which enables the model to operate within structured multi-agent frameworks.

\subsubsection{GUI Knowledge Injection}

\textbf{QA \& VQA.}
In addition to trajectory-format data, we further augment the agent's GUI knowledge through data in other modalities. As illustrated in Figure \ref{fig:capability_enhancement}, we crawl a substantial volume of data from diverse sources on the Internet to construct a knowledge base of GUI-related information, encompassing software feature configurations, operational instructions, website navigation, among others. The primary sources can be categorized into three types: (i) official documentation and tutorials of software applications (\textit{e.g.}, Microsoft Office, LibreOffice); (ii) software forums (\textit{e.g.}, WPS Academy) and Q\&A platforms (\textit{e.g.}, Baidu Jingyan); and (iii) web navigation information extracted from existing open-source web datasets. After data cleaning, the crawled information is rewritten by LLMs into task-level QA or step-level VQA data, thereby enhancing the agent's GUI knowledge.

\textbf{World Modeling.}
A capable GUI agent should not only perceive the current screen state but also anticipate how the interface will change in response to its actions. To cultivate this predictive understanding, we construct world modeling data derived from trajectory recordings. Specifically, given a screenshot and the action executed at that step, we prompt a proprietary model (\textit{e.g.}, Claude-4.5) to produce a fine-grained description of the subsequent screenshot, explicitly highlighting the state transitions.
For example, newly appeared dialogs, changed text fields, shifted focus, or updated visual elements.
These action-conditioned state-transition descriptions are then used as training supervision. Through this process, the model acquires an internalized understanding of GUI environment dynamics, enabling it to better reason about the consequences of candidate actions before execution, which in turn facilitates more informed decision-making in multi-step tasks.

\subsubsection{Unified CoT Synthesis}
% \paragraph{Unified CoT synthesis for all trajectories.}
% {\color{red}Finally, every trajectory (automated, human, or simulated)} is augmented with step-wise \emph{thought} and \emph{conclusion} through a unified CoT synthesis pipeline. For each step \(t\), we form the generation context
% \[
% x_t=\Big(\mathcal{I},\{(o_i,a_i)\}_{i<t},o_{t-1},o_t,a_t\Big),
% \]
% and generate
% \[
% \text{thought}_t=g_{\theta}(x_t),\qquad \text{conclusion}_t=h_{\theta}(x_t),
% \]
% where \(g_{\theta}\) produces a rationale grounded in the task and UI changes, and \(h_{\theta}\) outputs a concise decision summary (e.g., intended target, tool arguments, expected outcome). The final augmented trajectory is
% \[
% \hat{\tau}=\{(o_t,a_t,\text{thought}_t,\text{conclusion}_t)\}_{t=1}^{T}.
% \]
% This unified augmentation yields consistent reasoning supervision across devices and data sources, aligning trajectory learning with the instruct/thinking variants of GUI-Owl~1.5.

After obtaining trajectory data containing action sequences through various approaches (\textit{i.e.}, agent exploration, human annotation and virtual environments), we design a chain-of-thought (CoT) synthesis pipeline to generate corresponding thoughts and conclusions for each step in the trajectory data, thereby enhancing the agent's capabilities in screen observation, memory management, progress reflection, and tool invocation.

As illustrated in figure~\ref{fig:capability_enhancement}, given the $i$-th step of a trajectory, we first employ a vision-language model (VLM) to describe the screen content and further extract query-relevant information from it. For queries that require memorizing key on-screen information, such as \textit{Check the weather in Paris and London for next Monday and record it in the memo}, we extract information from the query-relevant content that may be needed in subsequent steps and incorporate it into the memory.
Furthermore, we feed the action parameters executed at the $i$-th step, the screenshots captured before and after execution, and the user query into the VLM to determine whether the execution outcome of this step aligns with expectations. If the change in screen state is consistent with expectations, the progress of the current task is updated accordingly in the subsequent step; otherwise, corresponding reflections and error corrections are generated to inform the next action decision.

The observation, memory, reflection and task progress information obtained above are then fed into an LLM to synthesize the thought and conclusion corresponding to the action at this step. Specifically, the thought simulates the agent's reasoning process of integrating these pieces of information for action decision-making, while the conclusion provides a concise action decision.
Moreover, if the current trajectory involves tool invocation, the tool definitions from the tool set are also provided as input to the LLM, so that the synthesized thought incorporates reasoning about tool selection and invocation.

The CoT synthesis pipeline designed above enables the model to: (i) reflect on the execution outcome of the previous action and analyze the overall task progress accordingly, thereby achieving superior long-horizon decision-making capability; and (ii) simultaneously record key on-screen information (\textit{e.g.}, prices, weather conditions) that may be required in subsequent steps during the operational process, thereby achieving enhanced memory capability.

\subsubsection{Multi-agent Collaboration}

To enable the model to serve not only as an end-to-end agent but also as the components within multi-agent frameworks for multi-agent collaboration, we additionally employ the Mobile-Agent-v3.5 framework for agent exploration during the trajectory collection phase.
Mobile-Agent-v3.5 agent framework largely follows Mobile-Agent-v3, and we only summarize the key components and interfaces here for completeness. The system instantiates a small set of role-specialized modules and executes them in a closed loop over a device environment (mobile/desktop/web), with a unified action abstraction and shared model backbone.

\paragraph{Problem setup.}
Given a user instruction \(I\) and the current device state \(S_t\) (e.g., screenshot, UI tree, device metadata), the goal is to produce an action \(a_t\in\mathcal{A}\) that drives the environment to \(S_{t+1}\sim P(\cdot\mid S_t,a_t)\) until termination.

\paragraph{Roles and state variables.}
We maintain four agent roles: a Manager (planner), a Worker (executor), a Reflector (verifier), and a Notetaker (memory). At step \(t\), the system state is summarized by
\[
X_t \triangleq (I, S_t, SS_t, F_{t-1}, N_t),
\]
where \(SS_t\) is the (ordered) subgoal list, \(F_{t-1}\) is the latest feedback, and \(N_t\) is persistent notes.

\paragraph{Manager: subgoal planning and update.}
The Manager decomposes the instruction into subgoals and dynamically updates them:
\[
SS_0 = f_M(I, K_{\text{RAG}}), \qquad SS_{t+1} = u_M(SS_t, F_t, S_{t+1}),
\]
where \(K_{\text{RAG}}\) denotes optionally retrieved external knowledge.

\paragraph{Worker: action generation.}
Given the current context, the Worker selects a subgoal and produces the next action (optionally as a structured tuple with rationale and a normalized action schema):
\[
a_t \sim \pi_W(\cdot \mid I, S_t, SS_t, F_{t-1}, N_t).
\]

\paragraph{Reflector: transition-level verification and feedback.}
After executing \(a_t\) on the device, the Reflector judges the transition and provides diagnostic feedback:
\[
(j_t, \phi_t) = f_R(S_t, a_t, S_{t+1}), \qquad j_t \in \{\text{SUCCESS}, \text{FAILURE}\},
\]
and we set \(F_t \triangleq (j_t,\phi_t)\).

\paragraph{Notetaker: persistent memory update.}
Upon successful progress, the Notetaker extracts and stores salient transient information for future steps:
\[
N_{t+1} =
\begin{cases}
u_C(N_t, S_{t+1}) & \text{if } j_t=\text{SUCCESS},\\
N_t & \text{otherwise}.
\end{cases}
\]

\paragraph{Execution loop.}
The framework iterates \((SS_t, a_t, F_t, N_t)\) updates until all subgoals are completed or a termination condition is met (e.g., success, timeout, or safety stop). This design isolates planning, execution, verification, and memory, while remaining compatible with the multi-platform interfaces used throughout training and evaluation.

\subsection{Training Paradigm}
GUI-Owl-1.5 is initialized from Qwen3-VL and trained through a three-stage process. Compared with GUI-Owl, each stage is substantially expanded in data diversity and task coverage to support multi-platform automation, tool invocation, and complex agentic interactions.

% \begin{itemize}
% \item Pre-training Phase:
% \item Iterative Tuning Phase:
% \item 
% \end{itemize}
\subsubsection{Pre-training}
We construct a large-scale pre-training corpus that extends beyond basic GUI understanding. In addition to the UI recognition and trajectory data used in GUI-Owl, we incorporate (i) QA and VQA knowledge data to strengthen general visual reasoning and knowledge comprehension, (ii) world-modeling data to train the model to predict how GUI states transition in response to actions, and (iii) tool invocation data to familiarize the model with tool-calling and MCP semantics from the earliest stage.

\subsubsection{Supervised Fine-tuning}
We perform supervised fine-tuning (SFT) to align GUI-Owl-1.5 with diverse agentic tasks across multiple devices. The SFT data covers multi-device trajectory data with CoT annotations (Section 2.2.1), augmented grounding data (Section 2.2.2), structured tool invocation supervision for both conventional tool calls and MCP-based interactions, and dedicated browser interaction data capturing the unique characteristics of web-based GUIs. This stage transforms the pre-trained model into a capable multi-device agent supporting GUI manipulation, tool invocation, and browser automation with explicit reasoning.

\subsubsection{Reinforcement Learning}

\begin{figure}[!t]
    \centering
    \includegraphics[width=0.99\textwidth]{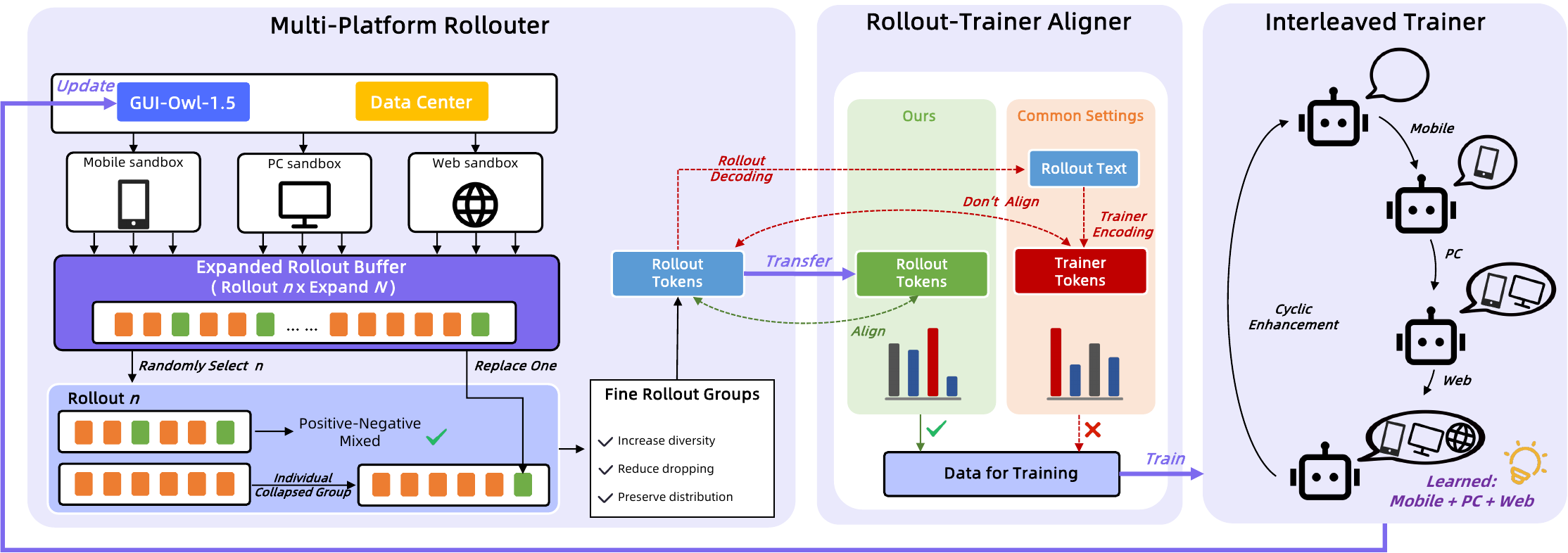}
    \vspace{-0.1cm}
    \caption{Overview of our reinforcement learning pipeline.}
    \label{fig:grounding_pipeline}
    \vspace{0.3cm}
\end{figure}

We perform a large-scale reinforcement learning called MRPO (Multi-platform Reinforcement Policy Optimization) to further align GUI-Owl-1.5 with long-horizon, tool-augmented GUI control across heterogeneous devices. The key challenges are: (i) unifying learning across mobile/desktop/web environments under one policy, (ii) stabilizing GRPO training when grouped rollouts collapse to identical outcomes. We address these issues as follows, (iii) ensuring log-probability consistency between environment-side inference and training-side optimization, and (iv) mitigating cross-device optimization interference.

\paragraph{Multi-device RL with a unified policy.}
We optimize a single policy \(\pi_\theta(a\mid o)\) over trajectories collected from multiple device families \(d\in\mathcal{D}=\{\text{mobile},\text{desktop},\text{web}\}\). Each device defines its own environment \(\mathcal{E}_d\), action space \(\mathcal{A}_d\), and observation stream. We model device heterogeneity via a device-conditioned policy:
\[
\pi_\theta(a\mid o,d),\qquad a\in\mathcal{A}_d.
\]

\paragraph{Online rollout buffer for GRPO under outcome collapse.}
We use GRPO-style grouped rollouts. For a task \(x\), we sample a group of \(n\) trajectories \(\{\tau_i\}_{i=1}^{n}\). In practice, it is common that all \(n\) rollouts yield identical terminal outcome (e.g., all success or all failure), making the group uninformative and often discarded. A replay buffer could increase diversity but introduces off-policy bias. We therefore propose an online rollout buffer that increases within-group diversity while remaining on-policy.

For each \(x\), we temporarily oversample \(kn\) rollouts on-policy:
\[
\mathcal{G}_{kn}(x)=\{\tau_i\}_{i=1}^{kn},\qquad \tau_i \sim \pi_\theta(\cdot\mid x),
\]
then uniformly subsample \(n\) trajectories to form the training group \(\mathcal{G}_n(x)\). Let \(Z(\tau)\in\{0,1\}\) denote a binary outcome (success/failure). The crucial property is that uniform subsampling preserves the marginal distribution of any statistic under on-policy sampling:
\[
\mathbb{E}\!\left[\frac{1}{n}\sum_{\tau\in \mathcal{G}_n(x)} f(\tau)\right]
=
\mathbb{E}_{\tau\sim\pi_\theta(\cdot\mid x)}[f(\tau)],
\]
because \(\mathcal{G}_{kn}(x)\) is i.i.d. on-policy and \(\mathcal{G}_n(x)\) is an exchangeable uniform subset. Thus oversample-then-subsample yields an approximately unbiased estimator.

Let \(Z(\tau)\in\{0,1\}\) be the terminal outcome (e.g., success). For a task \(x\), GRPO forms a group \(\mathcal{G}_n(x)=\{\tau_i\}_{i=1}^{n}\) with \(\tau_i\sim\pi_\theta(\cdot\mid x)\). The update becomes uninformative when the group is collapsed:
\[
\operatorname{Collapse}(\mathcal{G}_n)\ \triangleq\ \Big(\sum_{\tau\in\mathcal{G}_n} Z(\tau)\in\{0,n\}\Big).
\]
To reduce collapsed groups without introducing off-policy replay, we use an online oversample-and-select buffer. First, sample an on-policy pool of size \(kn\):
\[
\mathcal{G}_{kn}(x)=\{\tau_i\}_{i=1}^{kn},\qquad \tau_i\sim\pi_\theta(\cdot\mid x).
\]
Define the pool-diversity event
\[
\mathcal{A}\ \triangleq\ \Big(0<\sum_{\tau\in\mathcal{G}_{kn}} Z(\tau)<kn\Big),
\]
with probability
\[
\mathbb{P}(\mathcal{A}) \;=\; 1 - p^{kn} - (1-p)^{kn},\qquad
p\triangleq \mathbb{P}_{\tau\sim\pi_\theta(\cdot\mid x)}[Z(\tau)=1].
\]
Then construct the training group \(\widehat{\mathcal{G}}_n(x)\) by:
\[
\widehat{\mathcal{G}}_n(x)=
\begin{cases}
\operatorname{Subsample}_n(\mathcal{G}_{kn}(x)), & \neg \operatorname{Collapse}(\operatorname{Subsample}_n(\mathcal{G}_{kn}(x))),\\[2mm]
\operatorname{Swap1}\big(\operatorname{Subsample}_n(\mathcal{G}_{kn}(x)),\,\mathcal{G}_{kn}(x)\big), & \operatorname{Collapse}(\operatorname{Subsample}_n(\mathcal{G}_{kn}(x)))\ \wedge\ \mathcal{A},\\[2mm]
\varnothing, & \neg \mathcal{A},
\end{cases}
\]
where \(\operatorname{Subsample}_n(\cdot)\) is uniform random downsampling, and \(\operatorname{Swap1}(S,P)\) replaces one random element in \(S\) with a random opposite-outcome element from pool \(P\) (guaranteeing \(0<\sum_{\tau\in \widehat{\mathcal{G}}_n} Z(\tau)<n\)). This keeps all candidates strictly on-policy (sampled from current \(\pi_\theta\)) while sharply increasing the probability of obtaining a non-collapsed GRPO group.

\paragraph{Training--inference log-prob alignment via token-id transport.}
Our inference service is deployed on the environment side and returns trajectories as text (e.g., tool calls, typed strings, serialized actions). However, if the training-side tokenizer maps the returned text to token IDs differently from the inference-side tokenizer (due to non-unique segmentation), then the computed log-probabilities can be inconsistent:
\[
\log \pi_\theta(y\mid x)\Big|_{\text{train-tokenize}(y)}\ \neq\ \log \pi_\theta(y\mid x)\Big|_{\text{infer-tokenize}(y)}.
\]
This breaks KL regularization and policy-gradient estimators that assume the same sampled action representation. Our fix is to \emph{transport the original inference token IDs} alongside the textual payload. Concretely, for each generated sequence \(y\), the environment returns \((y,\, \mathbf{t}^{\text{infer}})\) where \(\mathbf{t}^{\text{infer}}=(t_1,\dots,t_L)\) are the exact token IDs used to sample \(y\). The training process then computes:
\[
\log \pi_\theta(y\mid x)\ :=\ \sum_{i=1}^{L}\log \pi_\theta\!\left(t_i\mid x, t_{<i}\right),
\]
thereby guaranteeing that the log-prob is evaluated on the same discrete event that was executed in the environment.

\paragraph{Alternating multi-device optimization to reduce gradient interference.}
Mixing trajectories from different devices in a single batch can induce strong gradient conflicts because \(\mathcal{A}_d\), UI conventions, and domain priors differ substantially. Let \(g_d=\mathbb{E}_{\tau\sim\pi_\theta,\mathcal{E}_d}[\nabla_\theta \mathcal{L}(\tau)]\) be the device-specific policy gradient. A naive mixture update uses \(g=\sum_d \lambda_d g_d\), but when \(\langle g_{d_1},g_{d_2}\rangle<0\) frequently, optimization becomes a “tug-of-war”. We adopt an alternating schedule across stages:
\[
\theta^{(s+1)} \leftarrow \theta^{(s)} - \eta \, g_{d_s},\qquad d_s \in \mathcal{D},
\]
where each stage \(s\) trains on a single device family (potentially with multiple environments within the family), and device families are visited cyclically or via a curriculum. This isolates device-specific adaptation while keeping a shared backbone, empirically improving stability and preserving cross-device generalization.

\begin{table}[!t]
\centering
\resizebox{\textwidth}{!}{%
\begin{tabular}{lccccc}
\toprule
\textbf{Agent Model} & \textbf{OSWorld-Verified} & \textbf{AndroidWorld} & \textbf{OSWorld-MCP} & \textbf{Mobile-World} & \textbf{WindowsAA} \\
\midrule
\rowcolor{gray!15}
\multicolumn{6}{l}{\textit{General Models}} \\
\midrule
SeedVL-1.5~\citep{seed2025seed1_5vl} & 34.1 & 62.1 & 38.4 & - & 39.6 \\
Claude-4-sonnet~\citep{claude4} & 43.9 & - & 43.3 & - & -\\
Claude-4-5-sonnet~\citep{claude45} & 62.9 & 56.0 & - & & -\\
Gemini-2.5-pro~\citep{gemini25} & - & 69.7 & 27.2 & - & -\\
Kimi K2.5~\citep{team2025kimi} & 63.3 & - & - & & -\\
Seed-1.8~\citep{seed18} & 61.9 & 70.7 & - & & - \\
OpenAI CUA o3~\citep{opencua} & 31.3 & - & - & & - \\
Qwen3-VL-8B-Instruct~\citep{qwen3_technical_report} & 33.9 & 47.6 & - & 5.5 & 28.8 \\
Qwen3-VL-8B-Think~\citep{qwen3_technical_report} & 33.9 & 50.0 & - & - & 24.1 \\
Qwen3-VL-32B-Instruct~\citep{qwen3_technical_report} & 32.6 & 57.3 & - & 9.0 & 30.9 \\
Qwen3-VL-32B-Think~\citep{qwen3_technical_report} & 41.0 & 63.7 & - & - & 42.9 \\
Qwen3-VL-235B-A22B-Instruct~\citep{qwen3_technical_report} & 31.6 & 63.7 & - & 9.5 & 28.9 \\
Qwen3-VL-235B-A22B-Think~\citep{qwen3_technical_report} & 38.1 & 62.0 & 39.1 & - & 32.1 \\
\midrule
\rowcolor{gray!15}
\multicolumn{6}{l}{\textit{GUI Models (Single-Platform)}} \\
\midrule
OpenCUA-7B~\citep{opencua} & 28.2 & - & - & - & - \\
OpenCUA-32B~\citep{opencua} & 34.8 & - & - & & - \\
OpenCUA-72B~\citep{opencua} & 45.0 & - & - & & - \\
EvoCUA~\citep{xue2026evocua} & 56.7 & - & - & & - \\
MAI-UI-8b~\citep{zhou2025mai} & - & 70.7 & - & 24.9 & - \\
MAI-UI-32b~\citep{zhou2025mai} & - & 73.3 & - & 37.3 & - \\
MAI-UI-235b-A22b~\citep{zhou2025mai} & - & 76.7 & - & 41.7 & - \\
\midrule
\rowcolor{gray!15}
\multicolumn{6}{l}{\textit{GUI Models (Multi-Platform)}} \\
\midrule
UI-TARS-72B-DPO~\citep{ui-tars-seed} & 27.1 & 46.6 & - & & - \\
UI-TARS-1.5-7B~\citep{ui-tars-15-seed} & 27.4 & - & - & & 15.9 \\
UI-TARS-1.5~\citep{ui-tars-15-seed} & - & 64.2 & - & 20.9 & 42.1 \\
UI-TARS-2~\citep{ui-tars-2-seed} & 53.1 & 73.3 & - & & 50.6 \\
GELab-Zero-4B~\citep{yan2025step} & 31.9 & 63.9 & - & 10.9 & - \\
GELab-Zero-8B~\citep{yan2025step} & 40.2 & 67.7 & - & & - \\
GUI-Owl-7b~\citep{ye2025mobile} & 34.9 & 66.4 & - & 4.5 & - \\
GUI-Owl-32b~\citep{ye2025mobile} & - & - & - & 5.5 & - \\
\midrule
\rowcolor{gray!15}
\multicolumn{6}{l}{\textit{Ours (Multi-Platform)}} \\
\midrule
GUI-Owl-1.5-2B-Instruct & {43.5} & 67.9 & 33.0 & 31.3 & 25.78 \\
GUI-Owl-1.5-4B-Instruct & {48.2} & 69.8 & 31.7 & 32.3 & 29.44 \\
GUI-Owl-1.5-8B-Instruct & {52.3} & 69.0 & 41.8 & 41.8 & 31.66 \\
% osworld-mcp 44.6
GUI-Owl-1.5-8B-Thinking & {52.9} & \textbf{71.6} & 38.8 & 33.3 & 35.07 \\
GUI-Owl-1.5-32B-Instruct & \textbf{56.5} & 69.8 & \textbf{47.6} & \textbf{46.8} & 44.76 \\
GUI-Owl-1.5-32B-Thinking & 56.0 & 69.8 & 43.8 & 42.8 & 44.13 \\
\bottomrule
\end{tabular}
}
\caption{Comparison with state-of-the-art methods on online computer use and mobile use benchmarks.}
\label{tab:online_results}
\vspace{-5em}
\end{table}

\begin{table}[H]
    \centering
    \footnotesize
    \resizebox{\textwidth}{!}{
    \setlength{\tabcolsep}{4pt}
    \small
    \begin{tabular}{@{}lcccc@{}}
        \toprule
        \textbf{Agent Model} & \makecell[c]{\textbf{WebArena}} & \makecell[c]{\textbf{VisualWebArena}} & \makecell[c]{\textbf{WebVoyager}} & \makecell[c]{\textbf{Online-Mind2Web}} \\
        \midrule
        \rowcolor{gray!15}
        \multicolumn{5}{l}{\textit{Proprietary Models}} \\
        \midrule
        % WebVoyager + GPT-4~\citep{he2024webvoyager} & - & - & 59.1 & - \\
        % Agent-E + GPT-4-Turbo~\citep{abuelsaad2024agent} & - & - & 73.1 & 28.0 \\
        % Claude Computer Use 3.5~\citep{claude35} & - & - & - & 29.0 \\
        Browser-Use~\citep{browser-use} & - & - & 89.1 & 30.0 \\
        Claude-CUA-3.7~\citep{claude37} & - & - & - & 56.3 \\
        Operator~\citep{cua2025} & - & - & 87.0 & 61.3 \\
        Gemini-CUA~\citep{gemini3pro} & - & - & - & 69.0 \\
        Navigator~\citep{navigator} & - & - & - & 78.7 \\
        Magnitude + Claude-4-Sonnet~\citep{magnitude} & - & - & 93.9 & - \\
        VisualWebArena + GPT-4o~\citep{koh2024visualwebarena} & - & 19.8 & - & - \\
        Tree Search + GPT-4o~\citep{koh2024tree} & 19.2 & 26.4 & - & - \\
        % ExACT + GPT-4o~\citep{yu2024exact} & - & 33.7 & - & - \\
        % AWorld + GPT-5~\citep{AWorld} & - & 36.5 & - & - \\
        WALT + GPT-5~\citep{prabhu2025walt} & 50.1 & 52.9 & - & - \\
        SGV + Gemini-2.5-Flash~\citep{andrade2025let} & - & 54.4 & - & - \\
        % WebPilot + GPT-4o~\citep{zhang2025webpilot} & 37.2 & - & - & - \\
        % AgentOccam + GPT-4-Turbo~\citep{yang2024agentoccam} & 43.1 & - & - & - \\
        % Learn-by-Interact + Claude-3.5-Sonnet~\citep{su2025learn} & 48.0 & - & - & - \\
        % AgentSymbiotic + Claude-3.5~\citep{webtreesearch} & 52.1 & - & - & - \\
        DeepSky Agent + Claude-4-Sonnet~\citep{deepsky} & 66.9 & - & - & - \\
        OAgent + Gemini-3-Pro~\citep{oagent} & 71.6 & - & - & - \\
        \midrule
        \rowcolor{gray!15}
        \multicolumn{5}{l}{\textit{Open-Source Models}} \\
        \midrule
        WebStar-7B~\citep{he2026webstar} & - & - & 44.8 & 22.8 \\
        WebStar-32B~\citep{he2026webstar} & - & - & 48.6 & 23.8 \\
        DynaWeb-8B~\citep{ding2026dynaweb} & 31.0 & - & 38.7 & - \\
        % Qwen2-VL-7B~\citep{ICAL2025} & - & 2.9 & - & - \\
        % Qwen2-VL + ICAL~\citep{ICAL2025} & - & 8.2 & - & - \\
        ViGoRL-7B~\citep{ViGoRL} & - & 11.2 & - & - \\
        Llama-3-70B-Instruct + Tree Search~\citep{webtreesearch} & 10.1 & 16.7 & - & - \\
        AgentSymbiotic-8B~\citep{webtreesearch} & 43.2 & - & - & - \\
        \midrule
        \rowcolor{gray!15}
        \multicolumn{5}{l}{\textit{Ours}} \\
        \midrule
        \modelname-8B-Instruct  & 45.7 & 39.4 & 69.9 & 41.7 \\
        \modelname-8B-Thinking  & 46.7 & 40.8 & 78.1 & 48.6 \\
        \modelname-32B-Instruct & - & - & - & - \\
        \modelname-32B-Thinking & 48.4 & 46.6 & 82.1 & - \\
        \bottomrule
        
    \end{tabular}
    }
    \caption{Comparison with state-of-the-art methods on online browser use benchmarks.}
    \label{tab:web_datasets_comparison}    
    % \vspace{-3em}
\end{table}

\section{Experiments}
\subsection{Experimental Setup}
In this section, we evaluate GUI-Owl-1.5 across a wide range of benchmarks to thoroughly assess its performance as a native GUI agent for multi-device automation. Built on Qwen3-VL, GUI-Owl-1.5 comprises a family of models including instruct and thinking variants. In this report, we focus on 6 representative versions: GUI-Owl-1.5-2B-Instruct, GUI-Owl-1.5-4B-Instruct, GUI-Owl-1.5-8B-Instruct, GUI-Owl-1.5-8B-Think, GUI-Owl-1.5-32B-Instruct, and GUI-Owl-1.5-32B-Think. We conduct extensive experiments to evaluate GUI-Owl-1.5 along four key dimensions consistent with GUI-Owl: grounding capability, comprehensive GUI understanding, end-to-end agent capability, and multi-agent capability.

\subsection{Main Results}

\subsubsection{End2end and Multi-Agent capability on Online environment}

The benchmarks discussed above evaluate isolated, single-step actions, offering only a partial view of an agent's true capability. In practice, GUI automation requires chaining numerous decisions where earlier mistakes propagate and compound, and multiple valid execution paths may exist for the same task, yet offline benchmarks typically score against a single reference trajectory. To overcome these limitations, we conduct end-to-end evaluations across live interactive environments spanning three domains in Fig.~\ref{tab:online_results} and Fig.~\ref{tab:mem_benchmark}: Mobile Use (AndroidWorld~\citep{rawles2024androidworld}, MobileWorld~\citep{kong2025mobileworld}, and MMGUI-Bench~\citep{liu2026memguibench}), Computer Use (OSWorld~\citep{xie2024osworld}, WindowsAgentArena~\citep{bonatti2024windows}, and OSWorld-MCP~\citep{jia2025osworld}), and Browser Use (WebArena~\citep{zhou2023webarena}, VisualWebArena~\citep{koh2024visualwebarena}, WebVoyager~\citep{he2024webvoyager}, and Online-Mind2Web~\citep{xue2025an}). Among these, MMGUI-Bench specifically evaluates the agent's memory ability. MobileWorld and OSWorld-MCP further incorporate tool invocation, assessing the agent's ability to coordinate GUI operations with external tool and MCP calls. In all environments, success is determined solely by whether the final goal state is achieved, regardless of the specific path taken.

\textbf{Computer and Mobile Use}. As shown in Table~\ref{tab:online_results}, GUI-Owl-1.5 achieves state-of-the-art performance among multi-platform GUI models across both computer and mobile use benchmarks. On OSWorld-Verified, the most widely adopted computer use benchmark, GUI-Owl-1.5-8B-Thinking achieves 52.9, surpassing UI-TARS-2 (53.1) at comparable scale and outperforming all general-purpose models including Qwen3-VL-235B-A22B-Think (38.1). Even our 2B variant attains 43.5, exceeding models with over 10$\times$ more parameters such as UI-TARS-72B-DPO (27.1), showcasing strong parameter efficiency. Similarly, on WindowsAA, the 32B-Instruct model scores 44.76, outperforming all general-purpose models at comparable or larger scale. For mobile use, on AndroidWorld, our 8B-Thinking variant attains 71.6, on par with UI-TARS-2 (73.3). Beyond standard GUI interaction, Mobile-World and OSWorld-MCP further require coordinating GUI actions with external tool and MCP calls; on these two benchmarks, GUI-Owl-1.5-32B-Instruct scores 46.8 and 47.6 respectively, surpassing both single-platform specialists (\textit{e.g.}, MAI-UI-235B-A22B at 41.7) and leading proprietary models (\textit{e.g.}, Claude-4-Sonnet at 43.3 on OSWorld-MCP), demonstrating strong tool-use capability.

\textbf{Browser Use}. As shown in Table~\ref{tab:web_datasets_comparison}, GUI-Owl-1.5-8B-Thinking achieves 46.7 on WebArena, 40.8 on VisualWebArena, 78.1 on WebVoyager, and 48.6 on Online-Mind2Web, surpassing all open-source models by a wide margin and remaining competitive with proprietary systems. These results establish GUI-Owl-1.5 as one of the strongest open-source browser agents to date. Across almost all domains, the Thinking variants consistently outperform their Instruct counterparts, with pronounced gains on tasks requiring long-horizon planning (\textit{e.g.}, WebVoyager: 69.9 to 82.1, Online-Mind2Web: 41.7 to 48.6), validating the effectiveness of our thinking-mode training.

% \begin{quote}
% *A variant of GUI-Owl specifically RL-tuned for a desktop environment (Section 5.2). The general version of GUI-Owl achieves a score of 29.4.
% \end{quote}

\subsubsection{Grounding Capability}
The grounding capability evaluates a model’s ability to locate the corresponding UI element given a natural-language query. We use ScreenSpot Pro, OSWorld-G, OSWorld-G-Refine, ScreenSpot V2 and MMBench-GUI L2 as benchmarks. 
ScreenSpot V2 covers mobile, desktop, and web scenarios, while ScreenSpot-Pro primarily evaluates a model’s localization ability at ultra-high resolutions. OSWorld-G/OSWorld-G-Refine contains finely annotated queries. MMBench-GUI L2 has the broadest coverage and more faithfully reflects a model’s grounding performance in real-world settings. 
The performance comparisons are shown in \Cref{tab:mmbench_l2,tab:ssp,tab:osworld_g,tab:osworld_g_refine,tab:ssv2}.

In all grounding benchmarks, \modelname-32B-Instruct achieves state-of-the-art performance among all Multi-platform GUI Models.
Notably, on the Screenspot-Pro benchmark, which emphasizes high-resolution and challenging professional software grounding tasks, our \modelname-32B-Instruct achieves an accuracy of 72.9, surpassing all existing GUI agents (including single-platform, multi-platform, and grounding-specialized models) as well as the large-scale Gemini-3-Pro. 
Moreover, when augmented with a two-stage refinement strategy with crop tool—first localizing a coarse region, then cropping and zoomin for refined grounding, \modelname-32B-Instruct attains a substantially higher score of 80.3, outperforming all prior methods by a significant margin.

% MM_Bench_L2
% \vspace{-8em}
\begin{table}[H]
\centering

\resizebox{\textwidth}{!}{%
\begin{tabular}{l rrrrrrrrrrrr>{\columncolor{lightgray}}r}
\toprule
\multirow{2}{*}{\textbf{Model}} & \multicolumn{2}{c}{\textbf{Windows}} & \multicolumn{2}{c}{\textbf{MacOS}} & \multicolumn{2}{c}{\textbf{Linux}} & \multicolumn{2}{c}{\textbf{iOS}} & \multicolumn{2}{c}{\textbf{Android}} & \multicolumn{2}{c}{\textbf{Web}} & \cellcolor{white}\multirow{2}{*}{\textbf{Overall}} \\
\cmidrule(lr){2-3} \cmidrule(lr){4-5} \cmidrule(lr){6-7} \cmidrule(lr){8-9} \cmidrule(lr){10-11} \cmidrule(lr){12-13}
& Basic & Adv. & Basic & Adv. & Basic & Adv. & Basic & Adv. & Basic & Adv. & Basic & Adv. & \cellcolor{white}\\
\midrule
\rowcolor{gray!15}
\multicolumn{14}{l}{\textit{General Models}} \\
\midrule
GPT-4o~\citep{hurst2024gpt} & 1.48 & 1.10 & 8.69 & 4.34 & 1.05 & 1.02 & 5.10 & 3.33 & 2.53 & 1.41 & 3.23 & 2.92 & 2.87 \\
Claude-3.7~\citep{claude37}  & 1.48 & 0.74 & 12.46 & 7.51 & 1.05 & 0.00 & 13.69 & 10.61 & 1.40 & 1.40 & 3.23 & 2.27 & 4.66 \\
Qwen-2.5-Max-VL~\citep{Qwen2.5-VL} & 43.91 & 36.76 & 58.84 & 56.07 & 53.93 & 30.10 & 77.39 & 59.09 & 79.49 & 70.14 & 74.84 & 58.77 & 58.03 \\ 
InternVL3-72B~\citep{zhu2025internvl3} & 70.11 & 42.64 & 75.65 & 52.31 & 59.16 & 41.33 & 93.63 & 80.61 & 92.70 & 78.59 & 90.65 & 65.91 & 72.20 \\
\midrule
\rowcolor{gray!15}
\multicolumn{14}{l}{\textit{GUI Models (Single-Platform / Grounding Specialized)}} \\
\midrule
% ShowUI-2B~\citep{lin2025showui}  & 9.23 & 4.41 & 24.06 & 10.40 & 25.13 & 11.73 & 28.98 & 19.70 & 17.42 & 8.73 & 22.90 & 12.66 & 15.96 \\
UGround-V1-7B~\citep{gou2024navigating} & 66.79 & 38.97 & 71.30 & 48.55 & 56.54 & 31.12 & 92.68 & 70.91 & 93.54 & 70.99 & 88.71 & 64.61 & 65.68 \\
MAI-UI-8B~\citep{zhou2025mai} & 92.3 & 74.3 & 90.7 & 86.4 & 81.2 & 67.3 & 97.1 & 90.0 & 97.5 & 92.7 & 95.8 & 86.0 & 88.8 \\
MAI-UI-32B~\citep{zhou2025mai} & 93.0 &78.7 &92.8 &87.6 &86.9 &77.6 &97.1 &92.4 &98.0 &93.2 &96.1 &92.5 & 91.3\\
\midrule
\rowcolor{gray!15}
\multicolumn{14}{l}{\textit{GUI Models (Multi-Platform)}} \\
\midrule
Aguvis-7B-720P~\citep{xu2024aguvis}  & 37.27 & 21.69 & 48.12 & 33.27 & 33.51 & 25.00 & 67.52 & 65.15 & 60.96 & 50.99 & 61.61 & 45.45 & 45.66 \\
OS-Atlas-Base-7B~\citep{wu2024atlas}  & 36.90 & 18.75 & 44.35 & 21.68 & 31.41 & 13.27 & 74.84 & 48.79 & 69.60 & 46.76 & 61.29 & 35.39 & 41.42 \\
GUI-Owl-8B~\citep{ye2025mobile} & 86.35 & 61.76 & 81.74 & 64.45 & 74.35 & 61.73 & 94.90 & 83.03 & 95.78 & 83.66 & 93.22 & 72.72 & 80.49 \\
UI-TARS-1.5-7B~\citep{qin2025ui}  & 68.27 & 38.97 & 68.99 & 44.51 & 64.40 & 37.76 & 88.54 & 69.39 & 90.45 & 69.29 & 80.97 & 56.49 & 64.32 \\
UI-TARS-72B-DPO~\citep{qin2025ui}  & 78.60 & 51.84 & 80.29 & 62.72 & 68.59 & 51.53 & 90.76 & 81.21 & 92.98 & 80.00 & 88.06 & 68.51 & 74.25 \\ 
GUI-Owl-32B~\citep{ye2025mobile} & 85.61 & 65.07 & 84.93 & 67.05 & 76.96 & 63.27 & 95.22 & 85.45 & 96.07 & 87.04 & 95.48 & 80.84 & 82.97 \\
\midrule
\rowcolor{gray!15}
\multicolumn{14}{l}{\textit{Ours}} \\ \midrule
\modelname-2B-Instruct & 82.28 & 47.79 & 83.18 & 56.06 & 70.15 & 44.38 &88.53 & 69.39 & 81.69 & 69.10 & 90.32 & 70.12 & 72.17  \\
\modelname-4B-Instruct & 87.82 & 69.11 & 88.40 & 67.63 & 74.86 & 56.12 & 97.13 & 83.33 & 96.05 & 85.11 & 95.48 & 82.46 & 83.24 \\
\modelname-8B-Instruct & 89.66 & 65.44 & 88.11 & 72.83 & 72.77 & 56.63 & 95.85 & 83.93 & 95.21 & 82.86 & 93.22 & 77.59 & 82.52 \\
\modelname-8B-Thinking & 84.50 & 63.60 & 85.22 & 71.10 & 71.73 & 53.57 & 92.99 & 81.82 & 89.30 & 78.37 & 95.81 & 77.60 & 80.08 \\
\modelname-32B-Instruct & 91.51 & 68.75 & 92.46 & 77.46 & 76.44 & 67.35 & 97.13 & 90.61 & 96.06 & 89.04 & 96.13 & 84.74 & \textbf{86.84} \\
\modelname-32B-Thinking & 89.67 & 66.18 & 88.12 & 73.41 & 77.49 & 63.78 & 94.59 & 89.09 & 92.96 & 87.64 & 96.13 & 81.49 & 84.47 \\
\bottomrule
\end{tabular}%
}
\vspace{-0.8em}
\caption{
Comparison with state-of-the-art methods on the MMBench-GUI-L2 dataset.
% \underline{Underlined} denotes the second-best open-source performance.
}
\vspace{-3em}
\label{tab:mmbench_l2}
\end{table}
% SSP
\begin{table}[H]
\vspace{-0.8em}
\centering
% To make the table fit in the page width
\resizebox{\textwidth}{!}{%
\begin{tabular}{l rr rr rr rr rr rr >{\columncolor{lightgray}}r}
\toprule
\textbf{\multirow{2}{*}{Agent Model}}& \multicolumn{2}{c}{\textbf{Development}} & \multicolumn{2}{c}{\textbf{Creative}} & \multicolumn{2}{c}{\textbf{CAD}} & \multicolumn{2}{c}{\textbf{Scientific}} & \multicolumn{2}{c}{\textbf{Office}} & \multicolumn{2}{c}{\textbf{OS}} & \cellcolor{white}\textbf{\multirow{2}{*}{Avg}} \\
\cmidrule(lr){2-3} \cmidrule(lr){4-5} \cmidrule(lr){6-7} \cmidrule(lr){8-9} \cmidrule(lr){10-11} \cmidrule(lr){12-13}
 & \textbf{Text} & \textbf{Icon} & \textbf{Text} & \textbf{Icon} & \textbf{Text} & \textbf{Icon} & \textbf{Text} & \textbf{Icon} & \textbf{Text} & \textbf{Icon} & \textbf{Text} & \textbf{Icon} & \cellcolor{white}\\
\midrule
\rowcolor{gray!15}
\multicolumn{14}{l}{\textit{General Models}} \\
\midrule
% GPT-4o~\citep{hurst2024gpt} & 1.3 & 0.0 & 1.0 & 0.0 & 2.0 & 0.0 & 2.1 & 0.0 & 1.1 & 0.0 & 0.0 & 0.0 & 0.8 \\
Claude 3.7 Sonnet~\citep{claude37}  & - & - & - & - & - & - & - & - & - & - & - & - & 27.7 \\
Operator~\citep{cua2025}   & 50.0 & 19.3 & 51.5 & 23.1 & 16.8 & 14.1 & 58.3 & 24.5 & 60.5 & 28.3 & 34.6 & 30.3 & 36.6 \\
Gemini-3-Pro~\citep{gemini3pro} & - & - & - & - & - & - & - & - & - & - & - & - & 72.7 \\
Seed1.8~\citep{seed18} & - & - & - & - & - & - & - & - & - & - & - & - & $73.1^{\circ}$ \\
Qwen3-VL-8B-Instruct~\citep{Qwen3-VL} & - & - & - & - & - & - & - & - & - & - & - & - & 54.6  \\
Qwen3-VL-8B-Thinking~\citep{Qwen3-VL} & - & - & - & - & - & - & - & - & - & - & - & - & 46.6  \\
Qwen3-VL-32B-Instruct~\citep{Qwen3-VL} & - & - & - & - & - & - & - & - & - & - & - & - & 57.9  \\
Qwen3-VL-32B-Thinking~\citep{Qwen3-VL} & - & - & - & - & - & - & - & - & - & - & - & - & 57.1  \\
Qwen3-VL-235B-A22B-Instruct~\citep{Qwen3-VL} & - & - & - & - & - & - & - & - & - & - & - & - & 62.0  \\
Qwen3-VL-235B-A22B-Thinking~\citep{Qwen3-VL} & - & - & - & - & - & - & - & - & - & - & - & - & 61.8  \\
\midrule
\rowcolor{gray!15}
\multicolumn{14}{l}{\textit{GUI Models (Single-Platform / Grounding Specialized)}} \\
\midrule
% UI-R1-E-3B~\citep{lu2025ui}  &46.1& 6.9 &41.9 &4.2 &37.1 &12.5& 56.9& 21.8 &65.0 &26.4 &32.7& 10.1&33.5\\
% UI-TARS-7B~\citep{qin2025ui}  & 58.4 & 12.4 & 50.0 & 9.1 & 20.8 & 9.4 & 63.9 & 31.8 & 63.3 & 20.8 & 30.8 & 16.9 & 35.7 \\
InfiGUI-R1-3B~\citep{liu2025infigui}  & 51.3 & 12.4 & 44.9 & 7.0 & 33.0 & 14.1 & 58.3 & 20.0 & 65.5 & 28.3 & 43.9 & 12.4 & 35.7 \\
% JEDI-3B~\citep{xie2025scalingcomputerusegroundinguser}  & 61.0 & 13.8 & 53.5 & 8.4 & 27.4 & 9.4 & 54.2 & 18.2 & 64.4 & 32.1 & 38.3 & 9.0 & 36.1 \\
% GUI-G1-3B~\citep{zhou2025gui}  & 50.7 & 10.3 & 36.6 & 11.9 & 39.6 & 9.4 & 61.8 & 30.0 & 67.2 & 32.1 & 23.5 & 10.6 & 37.1 \\
JEDI-7B~\citep{xie2025scalingcomputerusegroundinguser}  & 42.9 & 11.0 & 50.0 & 11.9 & 38.0 & 14.1 & 72.9 & 25.5 & 75.1 & 47.2 & 33.6 & 16.9 & 39.5 \\
GUI-G$^{2}$-7B~\citep{tang2025guig2} &68.8& 17.2& 57.1& 15.4 &55.8 &12.5& 77.1& 24.5& 74.0 &32.7& 57.9& 21.3& 47.5 \\
OpenCUA-7B~\citep{opencua} & - & - & - & - & - & - & - & - & - & - & - & - & 50.0 \\
% SE-GUI-7B~\citep{yuan2025enhancing} & 68.2 & 19.3 & 57.6 & 9.1 & 51.3 & 42.2 & 75.0 & 28.2 & 78.5 & 43.4 & 49.5 & 25.8 & 47.3 \\
% GTA1-7B~\citep{yang2025gta1}  & 66.9 & 20.7 & 62.6 & 18.2 & 53.3 & 17.2 & 76.4 & 31.8 & 82.5 & 50.9 & 48.6 & 25.9 & 50.1 \\
GTA1-New-7B~\citep{yang2025gta1} & - & - & - & - & - & - & - & - & - & - & - & - & 55.5  \\
MAI-UI-8B~\citep{zhou2025mai} & 83.8 & 52.4 & 76.3 & 33.6 & 72.6 & 35.9 & 79.9 & 37.3 & 88.7 & 60.4 & 76.6 & 49.4 & 65.8 \\
MAI-UI-8B + Zoom-in~\citep{zhou2025mai} & 78.6 & 58.6 & 78.8 & 46.9 & 80.7 & 43.8 & 86.1 & 49.1 & 88.1 & 81.1 & 76.6 & 51.7 & $70.9^{\circ}$ \\
EvoCUA-8B~\citep{xue2026evocua} & - & - & - & - & - & - & - & - & - & - & - & - & 45.4  \\
UI-TARS-72B~\citep{qin2025ui} & 63.0 & 17.3 & 57.1 & 15.4 & 18.8 & 12.5 & 64.6 & 20.9 & 63.3 & 26.4 & 42.1 & 15.7 & 38.1 \\
UI-Venus-72B~\citep{ui-venus} & 84.4 & 33.1 & 73.2 & 30.8   & 66.5 & 29.7 & 84.7 & 42.7 & 83.1 & 60.4 & 75.7 & 36.0  & 61.9\\
UGround-v1-72B~\citep{uground}   & 55.8 & 4.8 & 54.0 & 10.5 & 16.8 & 4.7 & 70.8 & 22.7
            & 61.0 & 18.9 & 40.2 & 7.9  & 34.5 \\
GTA1-32B~\citep{yang2025gta1}  & 82.5 & 28.3 & 69.2 & 14.7 & 43.7 & 23.4 & 79.9 & 31.8 & 80.8 & 43.4 & 70.1 & 32.6 & 53.6 \\
GTA1-New-32B~\citep{yang2025gta1} & - & - & - & - & - & - & - & - & - & - & - & - & 63.6  \\
GTA1-72B~\citep{yang2025gta1}  & 79.9 & 33.1 & 73.2 & 20.3 & 56.9 & 28.1 & 81.9 & 38.2 & 85.3 & 49.1 & 73.8 & 37.1 & 58.4 \\
MAI-UI-32B~\citep{zhou2025mai} & 86.4 & 40.7 & 82.8 & 37.8 & 70.1 & 45.3 & 91.7 & 46.4 & 90.4 & 71.7 & 78.5 & 34.8 & 67.9 \\
MAI-UI-32B + Zoom-in~\citep{zhou2025mai} & 84.4 & 57.9 & 87.9 & 46.2 & 79.2 & 53.1 & 91.7 & 54.5 & 88.1 & 79.2 & 80.4 & 47.2 & $73.5^{\circ}$ \\
EvoCUA-32B~\citep{xue2026evocua} & - & - & - & - & - & - & - & - & - & - & - & - & 49.7  \\
\midrule
\rowcolor{gray!15}
\multicolumn{14}{l}{\textit{GUI Models (Multi-Platform)}} \\
\midrule
GUI-Owl-7B~\citep{ye2025mobile} & 76.6 & 31.0 & 59.6 & 27.3 & 64.5 & 21.9 & 79.1 & 37.3 & 77.4 & 39.6 & 59.8 & 33.7 & 54.9 \\
GUI-Owl-32B~\citep{ye2025mobile} & 84.4 & 39.3 & 65.2 & 18.2 & 62.4 & 28.1 & 82.6 & 39.1 & 81.4 & 39.6 & 70.1 & 36.0 & 58.0 \\
UI-TARS-1.5~\citep{ui-tars-15-seed} & - & - & - & - & - & - & - & - & - & - & - & - & 61.6 \\
% GTA1-7B~\citep{yang2025gta1}  & 66.9 & 20.7 & 62.6 & 18.2 & 53.3 & 17.2 & 76.4 & 31.8 & 82.5 & 50.9 & 48.6 & 25.9 & 50.1 \\
Step-GUI-8B~\citep{yan2025step} & - & - & - & - & - & - & - & - & - & - & - & - & 62.6  \\

\midrule
\rowcolor{gray!15}
\multicolumn{14}{l}{\textit{Ours}} \\
\midrule
\modelname-2B-Instruct & 74.0 & 43.4 & 66.1 & 32.8 & 53.2 & 26.5 & 78.4 & 39.0 & 79.6 & 45.2 & 66.3 & 50.5 & 57.8 \\
 \textit{+ Zoom-In} & 77.2 & 62.0 & 71.7 & 56.6 & 75.1 & 51.5 & 84.0 & 49.0 & 88.7 & 77.3 & 73.8 & 53.9 & $70.4^{\circ}$ \\
\modelname-4B-Instruct & 83.1 & 57.9 & 73.7 & 41.9 & 59.8 & 45.3 & 87.5 & 43.6 & 87.0 & 60.3 & 80.3 & 50.5 & 66.8 \\
 \textit{+ Zoom-In} & 88.9 & 58.6 & 79.7 & 57.3 & 83.7 & 56.2 & 90.2 & 53.6 & 90.3 & 77.3 & 81.3 & 62.9 & $75.6^{\circ}$ \\
\modelname-8B-Instruct & 87.0 & 63.4 & 79.2 & 45.4 & 76.1 & 43.7 & 87.5 & 47.2 & 89.2 & 56.6 & 82.2 & 49.4 & 71.1 \\
 \textit{+ Zoom-In} & 90.2 & 68.9 & 84.8 & 56.6 & 86.8 & 62.5 & 89.5 & 55.4 & 91.5 & 71.6 & 86.9 & 53.9 & $77.8^{\circ}$ \\
\modelname-8B-Thinking & 85.7 & 37.9 & 68.2 & 28.7 & 56.9 & 28.1 & 75.7 & 30.9 & 83.6 & 35.8 & 69.2 & 38.2 & 57.6 \\
 \textit{+ Zoom-In} & 88.3 & 58.6 & 78.8 & 45.5 & 84.3 & 48.4 & 86.8 & 45.5 & 91.0 & 64.2 & 84.1 & 52.8 & $72.5^{\circ}$ \\
\modelname-32B-Instruct & 88.3 & 64.1 & 78.8 & 44.1 & 80.2 & 48.4 & 90.3 & 54.5 & 91.5 & 56.6 & 86.9 & 44.9 & \textbf{72.9} \\
 \textit{+ Zoom-In} & 92.2 & 73.1 & 82.8 & 67.1 & 89.3 & 65.6 & 91.7 & 57.3 & 93.2 & 75.5 & 86.0 & 57.3 & $\textbf{80.3}^{\circ}$ \\
\modelname-32B-Thinking & 83.1 & 37.2 & 71.2 & 23.8 & 46.7 & 20.3 & 81.3 & 34.5 & 87.0 & 47.2 & 72.0 & 31.5 & 57.0 \\
 \textit{+ Zoom-In} & 88.3 & 53.1 & 81.3 & 45.5 & 83.8 & 42.2 & 87.5 & 46.4 & 92.7 & 66.0 & 83.2 & 55.1 & $72.4^{\circ}$ \\
\bottomrule
\end{tabular}%
}
\vspace{-0.8em}
\caption{
Comparison with state-of-the-art methods on the ScreenSpot-Pro dataset. 
% \underline{Underlined} denotes the second-best open-source performance.
Values marked with $^{\circ}$ were processed with the crop tool.
}
\label{tab:ssp}
\vspace{-0.8em}
\end{table}
% 

% OSG
\begin{table}[H]
    % \vspace{-1em}
    \centering
    \resizebox{\textwidth}{!}{%
    \begin{tabular}{@{}lccccc>{\columncolor{lightgray}}r}
        \toprule
        \textbf{Agent Model} & \makecell[c]{\textbf{Text}\\\textbf{Matching}} & \makecell[c]{\textbf{Element}\\\textbf{Recognition}} & \makecell[c]{\textbf{Layout}\\\textbf{Understanding}} & \makecell[c]{\textbf{Fine-grained}\\\textbf{Manipulation}}& \textbf{Refusal} & \cellcolor{white}\textbf{Avg} \\
        \midrule
        \rowcolor{gray!15}
        \multicolumn{7}{l}{\textit{General Models}} \\
        \midrule
        Operator~\citep{cua2025}  & 51.3 & 42.4 & 46.6 & 31.5 & 0.0 & 40.6 \\
        Seed1.5-VL~\citep{seed2025seed1_5vl} & 73.9 & 66.7 & 69.6 & 47.0 & 18.5 &62.9 \\
        Qwen3-VL-8B-Instruct~\citep{Qwen3-VL}&69.0 &55.5 &59.7 &47.7 &- &54.8\\
        Qwen3-VL-8B-Thinking~\citep{Qwen3-VL}&- &- &- &- &- &56.7\\
        Qwen3-VL-32B-Instruct~\citep{Qwen3-VL}&- &- &- &- &- &65.1\\
        Qwen3-VL-32B-Thinking~\citep{Qwen3-VL}&- &- &- &- &- &64.0\\
        \midrule
        \rowcolor{gray!15}
        \multicolumn{7}{l}{\textit{GUI Models (Single-Platform / Grounding Specialized)}} \\
        \midrule
        % JEDI-3B \citep{xie2025scalingcomputerusegroundinguser} & 67.4 & 53.0 & 53.8 & 44.3 & 7.4 & 50.9 \\
        
        UGround-7B \citep{gou2024navigating} & 51.3 & 40.3 & 43.5 & 24.8 & 0.0 & 36.4 \\
        Aguvis-7B \citep{xu2024aguvis} & 55.9 & 41.2 & 43.9 & 28.2 & 0.0 & 38.7 \\
        
        JEDI-7B \citep{xie2025scalingcomputerusegroundinguser} & 65.9 & 55.5 & 57.7 & 46.9 & 7.4 & 54.1 \\
        
        GTA1-7B~\citep{yang2025gta1} & 42.1 & 65.7 & 62.7 & 56.1 & 0.0 & 55.1 \\
        
        UI-Venus-7B~\citep{ui-venus} & 74.6 & 60.5 & 61.5 & 45.5 & - & 58.8 \\
        MAI-UI-8B~\citep{zhou2025mai} & 72.0 &63.3 &66.0 &51.0 &- & 60.1 \\
        OpenCUA-32B \citep{opencua} & - & - & - & - & - & 59.6 \\
        GTA1-32B~\citep{yang2025gta1} & 63.2 & 78.4 & 73.3 & 65.2 &0.0&65.2  \\
        MAI-UI-32B~\citep{zhou2025mai} & 73.6 &72.4& 73.9 &57.7 &- & 67.6 \\
        EvoCUA-32B~\citep{xue2026evocua} & - &- &- &- &- &63.9\\
        % UI-Venus-72B~\citep{ui-venus} & 82.1 & 71.2 & 70.7 & 64.4 & - & 70.4 \\
        \midrule
        \rowcolor{gray!15}
        \multicolumn{7}{l}{\textit{GUI Models (Multi-Platform)}} \\
        \midrule
        OS-Atlas-7B \citep{wu2024atlas}& 44.1 & 29.4 & 35.2 & 16.8 & 7.4 & 27.7 \\
        UI-TARS-72B \citep{qin2025ui}& 69.4 & 60.6 & 62.9 & 45.6 & 0.0 &57.1 \\
        UI-TARS-7B \citep{qin2025ui} & 60.2 & 51.8 & 54.9 & 35.6 & 0.0 & 47.5 \\
        UI-TARS-1.5-7B~\citep{ui-tars-15-seed} & 36.8 & 62.7 & 62.2 & 50.8 & 0.0 & 52.8\\
        GUI-Owl-7B~\citep{ye2025mobile} & 64.8 & 63.6 & 61.3 & 41.0 & - & 55.9 \\
        GUI-Owl-32B~\citep{ye2025mobile} & 67.0 & 64.5 & 67.2 & 45.6 & - & 58.0 \\
        % Qwen3-VL-235B-A22B-Instruct~\citep{Qwen3-VL} & - & - & - & - & - & 66.7 \\
        % Qwen3-VL-235B-A22B-Thinking~\citep{Qwen3-VL} & - & - & - & - & - & 68.3 \\
        \midrule
        \rowcolor{gray!15}
        \multicolumn{7}{l}{\textit{Ours}} \\
        \midrule
        \modelname-2B-Instruct & 49.3 & 52.7 & 48.7 & 52.4 & 53.7 & 52.8  \\
        \modelname-4B-Instruct & 66.1 & 66.1 & 64.8 & 64.8 & 35.2 & 63.7 \\
        \modelname-8B-Instruct & 67.8 & 68.5 & 68.5 & 65.5 & 42.6 & 65.8 \\
        \modelname-8B-Thinking & 56.9 & 60.0 & 60.7 & 53.8 & 22.2 & 55.0 \\
        \modelname-32B-Instruct & 68.2 & 66.8 & 65.5 & 62.1 & 70.4 & \textbf{66.8} \\
        \modelname-32B-Thinking & 63.6 & 63.9 & 67.0 & 55.2 & 5.6 & 57.6 \\

        \bottomrule
    \end{tabular}
    }
    \caption{
    Comparison with state-of-the-art methods on the OSWorld-G dataset. 
    % \underline{Underlined} denotes the second-best open-source performance.
    }
    \label{tab:osworld_g}    
    \vspace{-0.8em}
\end{table}

\begin{table}[H]

    \centering
    \setlength{\tabcolsep}{4pt}
    \small
    \begin{tabular}{lccccc>{\columncolor{lightgray}}r}
        \toprule
        \textbf{Agent Model} & \makecell[c]{\textbf{Text}\\\textbf{Matching}} & \makecell[c]{\textbf{Element}\\\textbf{Recognition}} & \makecell[c]{\textbf{Layout}\\\textbf{Understanding}} & \makecell[c]{\textbf{Fine-grained}\\\textbf{Manipulation}}& \textbf{Refusal} & \cellcolor{white}\textbf{Avg} \\
        \midrule
        \rowcolor{gray!15}
        \multicolumn{7}{l}{\textit{General Models}} \\
        \midrule
        Operator~\citep{cua2025} & - & - & - & - & - & 57.8 \\
        Qwen3-VL-8B-Instruct~\citep{Qwen3-VL} & 73.9 &68.2 &73.1 &54.4 &- &64.4\\
        Qwen3-VL-32B-Instruct~\citep{Qwen3-VL} & 77.4 &73.6 &76.3 &57.7 &- &69.0\\
        \midrule
        \rowcolor{gray!15}
        \multicolumn{7}{l}{\textit{GUI Models (Single-Platform / Grounding Specialized)}} \\
        \midrule
        JEDI-7B \citep{xie2025scalingcomputerusegroundinguser} & - & - & - & - & - & 63.8 \\
        GTA1-7B~\citep{yang2025gta1} & 63.2 & 82.1 & 74.2  & 70.5 & 0.0 & 67.7 \\
        % MAI-UI-8B~\citep{zhou2025mai} & 77.4 &73.0 &78.3 &55.7&- & 68.6\\
        % MAI-UI-32B~\citep{zhou2025mai} & \textbf{79.7} &79.4 &81.0 &61.7 &- & 73.9 \\
        OpenCUA-32B \citep{opencua} & 63.2 & 79.9 & 84.9 & 62.1 & 7.4 & 70.2 \\
        GTA1-32B~\citep{yang2025gta1} & 63.2  &  83.6 & 84.4 & 70.5 & 0.0 & 72.2 \\
        \midrule
        \rowcolor{gray!15}
        \multicolumn{7}{l}{\textit{GUI Models (Multi-Platform)}} \\
        \midrule
        UI-TARS-1.5-7B~\citep{seed2025seed1_5vl} & 52.6 & 75.4 & 72.4 & 66.7 & 0.0 & 64.2\\
        \midrule
        \rowcolor{gray!15}
        \multicolumn{7}{l}{\textit{Ours}} \\
        \midrule
        \modelname-2B-Instruct & 65.1 & 60.3 & 63.3 & 65.9 & 53.7 & 62.6 \\
        \modelname-4B-Instruct & 73.1 & 71.8 & 72.9 & 74.1 & 29.6 & 68.4 \\
        \modelname-8B-Instruct & 73.1 & 69.4 & 71.0 & 74.8 & 42.5 & 69.3 \\
        \modelname-8B-Thinking & 64.7 & 61.1 & 64.9 & 65.3 & 20.4 & 59.8 \\
        \modelname-32B-Instruct & 68.5 & 68.5 & 71.0 & 70.1 & 68.5 & \textbf{69.7} \\
        \modelname-32B-Thinking & 68.9 & 69.7 & 72.6 & 72.1 & 5.6 & 64.7 \\

        \bottomrule
    \end{tabular}
    \caption{
    Performance comparison of state-of-the-art models on the OSWorld-G-Refine. 
    % \underline{Underlined} denotes the second-best open-source performance.
    }
    \label{tab:osworld_g_refine}
\end{table}

% SPv2
\begin{table*}[h]
\centering
\setlength{\tabcolsep}{10pt} % Adjust column spacing for a better fit
\resizebox{0.9\textwidth}{!}{
\begin{tabular}{l rr rr rr >{\columncolor{lightgray}}r}
\toprule
\textbf{\multirow{2}{*}{Agent Model}} & \multicolumn{2}{c}{\textbf{Mobile}} & \multicolumn{2}{c}{\textbf{Desktop}} & \multicolumn{2}{c}{\textbf{Web}} & \cellcolor{white}\textbf{\multirow{2}{*}{Overall}} \\
\cmidrule(lr){2-3} \cmidrule(lr){4-5} \cmidrule(lr){6-7}
 & \textbf{Text} & \textbf{Icon} & \textbf{Text} & \textbf{Icon} & \textbf{Text} & \textbf{Icon} & \cellcolor{white} \\
\midrule
\rowcolor{gray!15}
\multicolumn{8}{l}{\textit{General Models}} \\ \midrule
OmniParser-v2~\citep{yu2025omniparser}        & 95.5 & 74.6 & 92.3 & 60.9 & 88.0 & 59.6 & 80.7 \\
Operator~\citep{cua2025}             & 47.3 & 41.5 & 90.2 & 80.3 & 92.8 & 84.3 & 70.5 \\
Claude 3.7 Sonnet~\citep{claude37}    & -    & -    & -    & -    & -    & -    & 87.6 \\
UI-TARS-1.5~\citep{qin2025ui}          & -    & -    & -    & -    & -    & -    & 94.2 \\
Seed-1.5-VL~\citep{seed2025seed1_5vl}         & -    & -    & -    & -    & -    & -    & 95.2 \\
Qwen3-VL-8B-Instruct~\citep{Qwen3-VL} & - & - & - & - & - & - & 94.4 \\
Qwen3-VL-8B-Thinking~\citep{Qwen3-VL} & - & - & - & - & - & - & 93.5 \\
Qwen3-VL-32B-Instruct~\citep{Qwen3-VL} & - & - & - & - & - & - & 95.8 \\
Qwen3-VL-32B-Thinking~\citep{Qwen3-VL} & - & - & - & - & - & - & 95.7 \\
\midrule
\rowcolor{gray!15}
\multicolumn{8}{l}{\textit{GUI Models (Single-Platform / Grounding Specialized)}} \\ \midrule
% SeeClick~\citep{cheng2024seeclick}              & 78.4 & 50.7 & 70.1 & 29.3 & 55.2 & 32.5 & 55.1 \\
JEDI-3B~\citep{xie2025scalingcomputerusegroundinguser}              & 96.6 & 81.5 & 96.9 & 78.6 & 88.5 & 83.7 & 88.6 \\
% Qwen2.5-VL-7B~\citep{Qwen2.5-VL}       & 97.6 & 87.2 & 90.2 & 74.2 & 93.2 & 81.3 & 88.8 \\
JEDI-7B~\citep{xie2025scalingcomputerusegroundinguser}              & 96.9 & 87.2 & 95.9 & 87.9 & 94.4 & 84.2 & 91.7 \\
GTA1-7B~\citep{yang2025gta1}                & 99.0 & 88.6 & 94.9 & 89.3 & 92.3 & 86.7 & 92.4 \\
GTA1-32B~\citep{yang2025gta1}                   & 98.6 & 89.1 & 96.4 & 86.4 & 95.7 & 88.7 & 93.2 \\
% GTA1-72B~\citep{yang2025gta1}                  & 99.3 & 92.4 & 97.4 & 89.3 & 95.3 & 91.6 & 94.8 \\
EvoCUA-32B~\citep{xue2026evocua} & - & - & - & - & - & - & 90.4 \\
UI-Venus-72B~\citep{ui-venus} & 99.7 & 93.8 & 95.9 & 90.0 & 96.2 & 92.6 & 95.3 \\
\midrule
\rowcolor{gray!15}
\multicolumn{8}{l}{\textit{GUI Models (Multi-Platform)}} \\ \midrule
% UI-TARS-2B~\citep{qin2025ui}           & 95.2 & 79.1 & 90.7 & 68.6 & 87.2 & 78.3 & 84.7 \\
OS-Atlas-Base-4B~\citep{wu2024atlas}     & 95.2 & 75.8 & 90.7 & 63.6 & 90.6 & 77.3 & 85.1 \\
OS-Atlas-Base-7B~\citep{wu2024atlas}    & 96.2 & 83.4 & 89.7 & 69.3 & 94.0 & 79.8 & 87.1 \\
UI-TARS-7B~\citep{qin2025ui}          & 96.9 & 89.1 & 95.4 & 85.0 & 93.6 & 85.2 & 91.6 \\
UI-TARS-72B~\citep{qin2025ui}         & 94.8 & 86.3 & 91.2 & 87.9 & 91.5 & 87.7 & 90.3 \\
GUI-Owl-7B~\citep{ye2025mobile} & 99.0 & 92.4 & 96.9 & 85.0 & 93.6 & 85.2 & 92.8 \\
GUI-Owl-32B~\citep{ye2025mobile} & 98.6 & 90.0 & 97.9 & 87.8 & 94.4 & 86.7 & 93.2 \\
\midrule
\rowcolor{gray!15}
\multicolumn{8}{l}{\textit{Ours}} \\ \midrule
\modelname-2B-Instruct & 92.9 & 83.1 & 94.8 & 86.4 & 90.1 & 87.6 & 89.7 \\
\modelname-4B-Instruct & 95.1 & 93.1 & 95.9 & 86.4 & 92.9 & 92.8 & 93.2 \\
\modelname-8B-Instruct & 97.4 & 90.5 & 96.4 & 90.7 & 94.2 & 89.7 & 93.7\\
\modelname-8B-Thinking & 95.8 & 90.5 & 97.4 & 90.0 & 95.0 & 87.7 & 93.2 \\
\modelname-32B-Instruct & 97.1 & 92.6 & 97.9 & 89.3 & 95.5 & 96.4 & \textbf{95.3} \\
\modelname-32B-Thinking & 96.5 & 90.5 & 96.9 & 86.4 & 93.8 & 90.8 & 93.2 \\

\bottomrule
\end{tabular}
}
\caption{
Comparison with state-of-the-art methods on the ScreenSpot-V2 dataset.
% \underline{Underlined} denotes the second-best open-source performance.
}
\label{tab:ssv2}
\end{table*}

% KB
\begin{table*}[h!]
\vspace{-1em}
\centering
\resizebox{\textwidth}{!}{%
\begin{tabular}{l rrr rrr rr>{\columncolor{lightgray}}r}
\toprule
& \multicolumn{3}{c}{\textbf{Interface Perception}} & \multicolumn{3}{c}{\textbf{Interaction Prediction}} & \multicolumn{2}{c}{\makecell{\textbf{Instruction}\\\textbf{Understanding}}}& \textbf{Avg} \\
\cmidrule(lr){2-4} \cmidrule(lr){5-7} \cmidrule(lr){8-9} \cmidrule(lr){10-10}
\textbf{Agent Model} & \textbf{state} & \textbf{widget} & \textbf{layout} & \textbf{effect} & \textbf{type} & \textbf{parameter} & \textbf{goal} & \textbf{plan} & \\
\midrule
\rowcolor{gray!15}
\multicolumn{10}{l}{\textit{Proprietary Models}} \\ \midrule
O3~\citep{openai2025o3} & 83.03 & 84.12 & 88.39 & 74.83 & 75.98 & 45.75 & 69.45 & 95.47 & 73.30\\
Gemini-2.5-Pro~\citep{comanici2025gemini}& 81.19 & 84.36 &87.10 & 71.03 & 73.25 & 46.97 & 67.72 & 92.56 & 71.69\\
GPT-5-Chat~\citep{openai2025gpt5}& 78.90 & 84.12 & 88.39 & 71.55 & 71.55 & 43.85 & 68.98 & 91.26 & 70.97 \\
Claude-Sonnet-4-5~\citep{anthropic2025claude45} & 74.77 & 81.52 & 82.58 & 49.83 & 70.19 & 43.33 & 70.30 & 91.56 & 66.53 \\
Doubao-V-Pro~\citep{seed2025seed1_5vl}& 72.48 & 83.65 & 81.29 & 67.24 & 75.64 & 41.07 & 33.07 & 94.17 & 63.42 \\
Claude-Sonnet-4~\citep{anthropic2025claude4}& 70.18 & 78.44 & 78.06 & 41.90 & 62.52 & 42.11 & 65.20 & 94.82 & 62.16 \\
\midrule
\rowcolor{gray!15}
\multicolumn{10}{l}{\textit{Open-Source Models}} \\ \midrule
Qwen3-VL-8B-Instruct~\citep{Qwen3-VL}       & 76.61 & 89.81 & 83.87 & 58.97 & 70.20  &   51.58    & 67.40 & 77.99 & 67.84 \\
Qwen3-VL-8B-Thinking~\citep{Qwen3-VL}       & 68.81 & 76.30 & 83.23 & 67.07 & 70.36 & 40.73 & 64.09 & 91.26 & 66.81 \\
Qwen2.5VL-72B~\citep{bai2025qwen2} & 69.27 & 77.49 & 80.00 & 61.72 & 64.91 & 38.99 & 62.20 & 85.44 & 63.88 \\
Qwen2.5VL-7B~\citep{bai2025qwen2}& 53.21 & 67.77 & 60.00 & 51.72 & 50.60 & 39.34 & 16.22 & 48.87 & 45.16 \\
UITARS-1.5-7B~\citep{ui-tars-15-seed}  & 49.54 & 59.48 & 59.35 & 22.24 & 59.11 & 34.32 & 38.74 & 55.34 & 44.27 \\
GUI-OWL-7B~\citep{ye2025mobile} & 60.09 & 64.93 & 63.23 & 21.55 & 55.37 & 36.05 & 21.26 & 39.81 & 40.74 \\
GLM-4.5~\citep{zeng2025glm} & 49.54 & 48.10 & 53.55 & 27.07 & 17.55 & 35.53 & 28.98 & 91.91 & 38.10 \\
\rowcolor{gray!15}
\multicolumn{10}{l}{\textit{Ours}} \\ \midrule
\modelname-2B-Instruct  & 60.09 & 77.73 & 72.26 & 44.48 & 47.04 & 41.95 & 62.99 & 47.57 & 54.12 \\
\modelname-4B-Instruct  & 75.23 & 88.15 & 82.58 & 55.69 & 65.02 & 54.22 & 69.45 & 70.55 & 66.64 \\
\modelname-8B-Instruct  & 77.98 & 88.86 & 84.52 & 66.90 & 71.92 & 61.61 & 73.54 & 80.58 & 72.90 \\

\modelname-8B-Thinking  & 75.69 & 90.05 & 87.74 & 67.41 & 68.23 & 53.43 & 67.72 & 77.67 & 69.60 \\

\modelname-32B-Instruct & 77.06 & 92.65 & 85.81 & 70.69 & 73.89 & 64.12 & 73.39 & 88.67 & \textbf{75.45} \\

\modelname-32B-Thinking  & 81.19 & 90.76 & 85.81 & 68.10 & 73.89 & 57.65 & 72.91 & 86.41 & 73.36 \\

\bottomrule
\end{tabular}%
}
\caption{
Comparison with state-of-the-art methods on the GUI Knowledge Benchmark.
% \underline{Underlined} denotes the second-best open-source performance.
}
\label{tab:kb}
\end{table*}
\begin{table}[h]
\footnotesize
\centering

\begin{tabular}{llc}
\toprule
\textbf{Agent Model} & \textbf{Type} & \textbf{Success Rate} \\
\midrule
\multicolumn{3}{l}{\textit{Proprietary / Workflow Models}} \\
\midrule
Agent-S2 w/ Gemini-2.5-Pro       & Workflow & \textbf{41.7} \\
M3A w/ Gemini-2.5-Pro            & Workflow & 39.6 \\
T3A w/ Gemini-2.5-Pro            & Workflow & 31.2 \\
Mobile-Agent-E w/ Gemini-2.5-Pro & Workflow & 12.5 \\
AppAgent w/ Gemini-2.5-Pro       & Workflow & 8.3 \\
Mobile-Agent-V2 w/ Gemini-2.5-Pro& Workflow & 8.3 \\
SeeAct w/ Gemini-2.5-Pro         & Workflow & 6.2 \\
\midrule
\multicolumn{3}{l}{\textit{Native Agent Models}} \\
\midrule
Qwen3-VL-8B-Instruct            & Model & 18.8 \\
GUI-Owl-7B                      & Model & 14.6 \\
UI-Venus-7B                     & Model & 14.6 \\
UI-TARS-1.5-7B                  & Model & 8.3 \\
CogAgent                        & Model & 0.0 \\
\midrule
GUI-Owl-1.5-8B                  & Model & 22.9 \\
GUI-Owl-1.5-32B                  & Model & 27.1 \\
\bottomrule
\end{tabular}
\caption{Evaluation results on MemGUI-Bench (Easy tasks).}
\label{tab:mem_benchmark}
\end{table}

\subsubsection{Comprehensive GUI Understanding}
\textbf{GUI Knowledge}. The GUI Knowledge Benchmark~\citep{shi2025gui} systematically evaluates whether a GUI model possesses sufficient knowledge across three dimensions: Interface Perception (state information understanding, widget function understanding, and layout semantics understanding), Interaction Prediction (action effect, action type prediction, and action parameter prediction), and Instruction Understanding (goal interpretation and task planning).
On this benchmark, \modelname-32B-Instruct achieves an overall accuracy of 75.45, establishing the highest performance among all evaluated models, including proprietary ones such as o3 (73.30)~\citep{openAI_o3_o4_mini} and Gemini-2.5-Pro (71.69)~\citep{gemini25}. It attains particularly strong results on widget function understanding and action parameter prediction, substantially outperforming all other models in these categories. 

\textbf{GUI Memory}. We further evaluate on MemGUI-Bench~\citep{liu2026memguibench} (Table~\ref{tab:mem_benchmark}), which assesses an agent's ability to recall and leverage interaction history over long horizons. Among native agent models, GUI-Owl-1.5-32B achieves 27.1, substantially outperforming all prior baselines including Qwen3-VL-8B-Instruct (18.8) and UI-TARS-1.5-7B (8.3). Even our 8B variant (22.9) surpasses all existing native baselines, confirming that our training recipe effectively instills long-horizon memory capabilities without relying on external workflow orchestration. 
% We note that the best workflow-based system, Agent-S2 w/ Gemini-2.5-Pro, achieves 41.7 by leveraging dedicated external memory modules, suggesting that integrating GUI-Owl-1.5 into such frameworks could yield further gains.

\subsection{Detailed Analyses}

\textbf{Effect of Virtual-enviroment trajectory Production and Unified CoT Synthesis}.
We conduct ablation experiments to validate two key components: virtual environment-based trajectory production (Table~\ref{tab:ablation_venv}) and unified CoT synthesis (Table~\ref{tab:ablation_cot}).

As shown in Table~\ref{tab:ablation_venv}, removing trajectory data produced by virtual environments leads to dramatic performance drops on both PC-Eval (75.4\% to 42.0\%) and Mobile-Eval (86.7\% to 50.0\%). Here, PC-Eval is an in-house benchmark focusing on atomic desktop operations such as drag and scroll, as well as office document and spreadsheet editing tasks; Mobile-Eval is an in-house benchmark covering popular Chinese mobile application scenarios including food delivery, ride-hailing, ticket booking, among others. The substantial degradation on both benchmarks confirms that our web-rendering-based virtual environments effectively bypass real-world exploration limitations—such as CAPTCHA interruptions and the lack of accurate feedback—and provide scalable, high-quality trajectories that are critical for mastering these challenging scenarios.

As shown in Table~\ref{tab:ablation_cot}, removing the unified CoT synthesis causes consistent drops on both OSWorld (52.9\% to 47.4\%) and AndroidWorld (71.6\% to 65.0\%), demonstrating that step-wise thought and conclusion augmentation provides essential reasoning supervision. By equipping each trajectory step with observation, memory, reflection, and progress tracking, CoT synthesis enables the model to plan over long horizons and retain key information across steps, which is particularly beneficial for multi-step online tasks across different platforms.

The two components are complementary: virtual environments improve trajectory \textit{coverage and quality}, while CoT synthesis enhances \textit{reasoning and decision-making supervision}.

% We conduct ablation experiments to validate two key components: virtual environment-based trajectory production and unified CoT synthesis. Results are reported in Table~\ref{tab:ablation}.

% Removing virtual environment data leads to consistent degradation, confirming that our web-rendering-based environments effectively bypass real-world exploration limitations (e.g., CAPTCHA interruptions, lack of accurate feedback) and provide scalable, high-quality trajectories. Removing the unified CoT synthesis also causes notable drops, demonstrating that step-wise thought and conclusion augmentation provides essential reasoning supervision across heterogeneous data sources and device types. The two components are complementary: virtual environments improve trajectory \textit{quality}, while CoT synthesis enhances \textit{reasoning supervision}.

% \begin{table}[t]
% \centering

% \begin{tabular}{lccc}
% \toprule
% \textbf{Configuration} & \textbf{OSWorld} & \textbf{AndroidWorld}\\
% \midrule
% Full Model              & [XX.X] & [XX.X] \\
% \quad w/o Virtual Env   & [XX.X] & [XX.X] \\
% \quad w/o CoT Synthesis & [XX.X] & [XX.X] \\
% \quad w/o Both          & [XX.X] & [XX.X] \\
% \bottomrule
% \end{tabular}
% \caption{Ablation study on GUI-Owl-1.5-8B-Instruct.}
% \label{tab:ablation}
% \end{table}

\begin{table}[t]
\centering

\begin{tabular}{ccc}
\toprule
\textbf{Unified CoT Synthesis} & \textbf{OSWorld} & \textbf{AndroidWorld}\\
\midrule
\quad \xmark & 47.4 & 65.0 \\
\quad \cmark & 52.9 & 71.6 \\
\bottomrule
\end{tabular}
\caption{Ablation study on the unified CoT synthesis pipeline. The experiments are conducted with GUI-Owl-1.5-8B-Thinking.}
\label{tab:ablation_cot}
\end{table}

\begin{table}[t]
\centering

\begin{tabular}{ccc}
\toprule
\textbf{Virtual Environments} & \textbf{PC-Eval} & \textbf{Mobile-Eval}\\
\midrule
\quad \xmark  & 42.0\% & 50.0\% \\
\quad \cmark & 75.4\% & 86.7\% \\
% 26 / 30 , 15/30
\bottomrule
\end{tabular}
\caption{Ablation study on the virtual environments. The experiments are conducted with GUI-Owl-1.5-8B-Thinking. PC-Eval is an in-house benchmark evaluating atomic operations such as drag and scroll, as well as office document and spreadsheet editing tasks. Mobile-Eval is an in-house benchmark evaluating popular Chinese mobile application scenarios, including food delivery, ride-hailing, ticket booking, among others.}
\label{tab:ablation_venv}
\end{table}

\textbf{Effect of Unstable-set Train and Interleaved Train in RL}.
\begin{figure}[!t]
    \centering
    \includegraphics[width=0.99\textwidth]{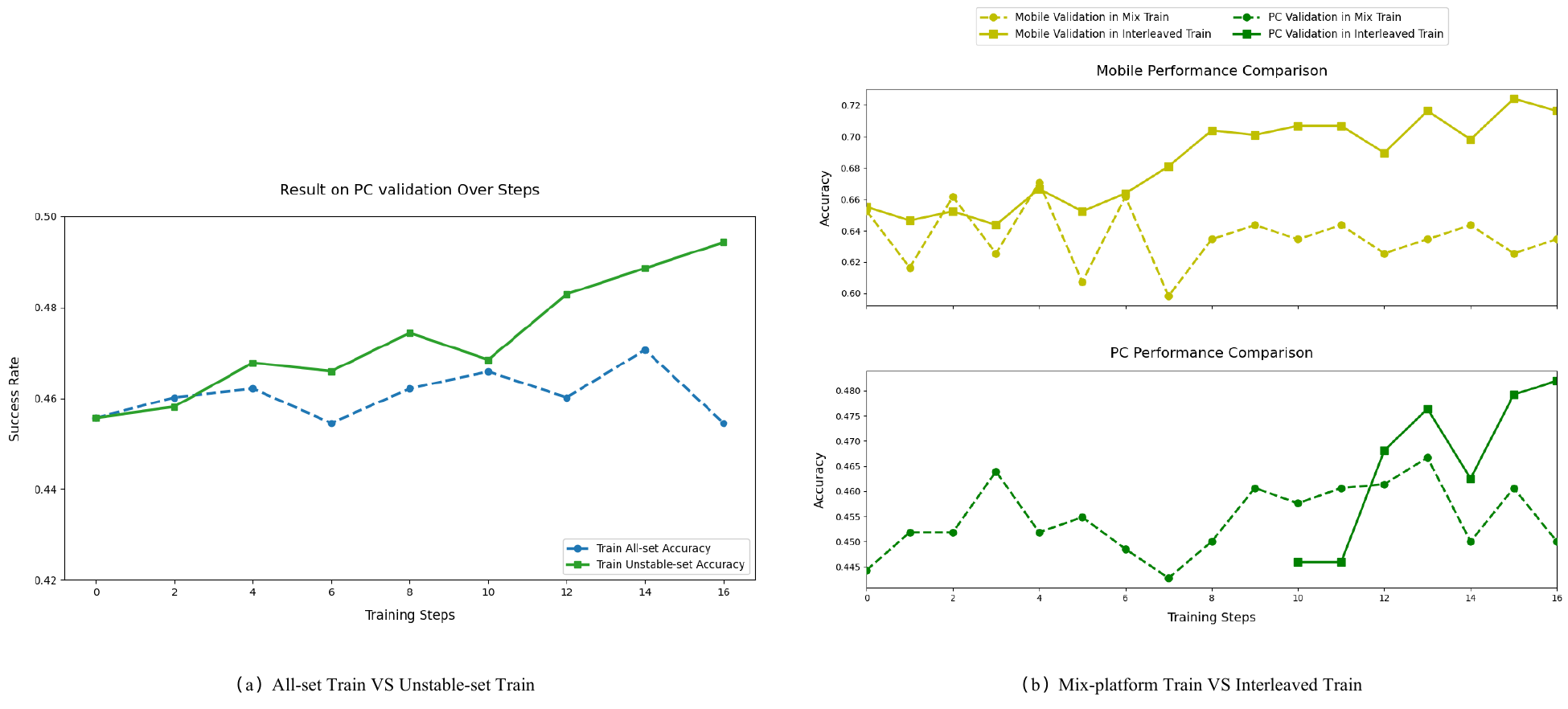}
    \vspace{-0.1cm}
    \caption{
    Ablation Study on Reinforcement Learning Training Strategies for \modelname-8B-thinking: Task Selection and Multi-Platform Training Strategies.
    }
    \label{fig:rl_ablation}
    \vspace{-0.25cm}
\end{figure}
We conduct ablation experiments to validate two critical training strategies for \modelname's reinforcement learning optimization, demonstrating the effectiveness of targeted task selection and multi-platform training strategy, as shown in \cref{fig:rl_ablation}.
\cref{fig:rl_ablation}(a) compares PC validation performance between full dataset training and unstable-task-only training (derived from multi-round rollouts).
Unstable-task-focused training achieves faster convergence and higher final accuracy, demonstrating the efficacy of prioritizing challenging tasks for robust model optimization. 
In \cref{fig:rl_ablation} (b), mix-platform training (simultaneous multi-platform data optimization) is contrasted with our interleaved training (switch from Mobile to PC at step 10). 
Mix-platform training exhibits performance oscillation due to cross-platform interference, whereas interleaved training enables focused optimization per platform while maintaining performance stability during transitions.
This approach achieves synergistic multi-platform growth, validating the superiority of our presented interleaved RL training strategy.

\subsection{Case Study}

\begin{figure}[!t]
    \centering
    \includegraphics[width=0.99\textwidth]{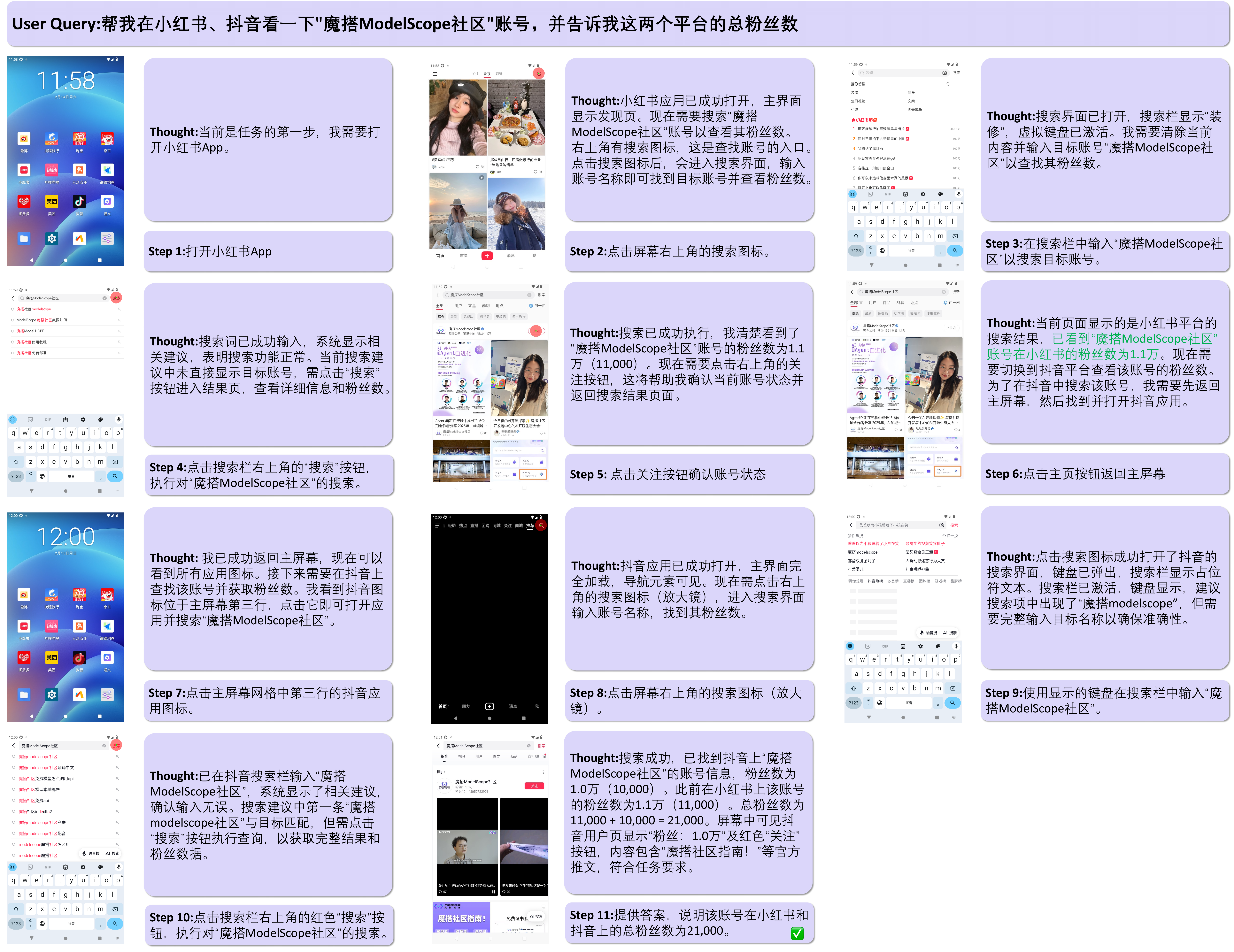}
    \vspace{-0.1cm}
    \caption{A complete operation process on the Android platform, in which the user query requires the agent to search and summarize information on social media platforms.}
    \label{fig:mobile_memory}
    \vspace{-0.25cm}
\end{figure}

We present three representative cases to illustrate the comprehensive capabilities of GUI-Owl 1.5 beyond basic GUI navigation.

\textbf{Mobile Use Case (\cref{fig:mobile_memory})} In this case, the user seeks to determine the total follower count of the ModelScope Community account across two social media platforms: Xiaohongshu and Douyin. The agent first launches the Xiaohongshu app, enters the account name in the search box, retrieves the follower information, and stores it in memory. Subsequently, the agent navigates to the Douyin app to obtain the corresponding follower count. By combining the retrieved information from memory with the current data, the agent calculates and reports the number of overall follower across both platforms.

\begin{figure}[!t]
    \centering
    \includegraphics[width=0.99\textwidth]{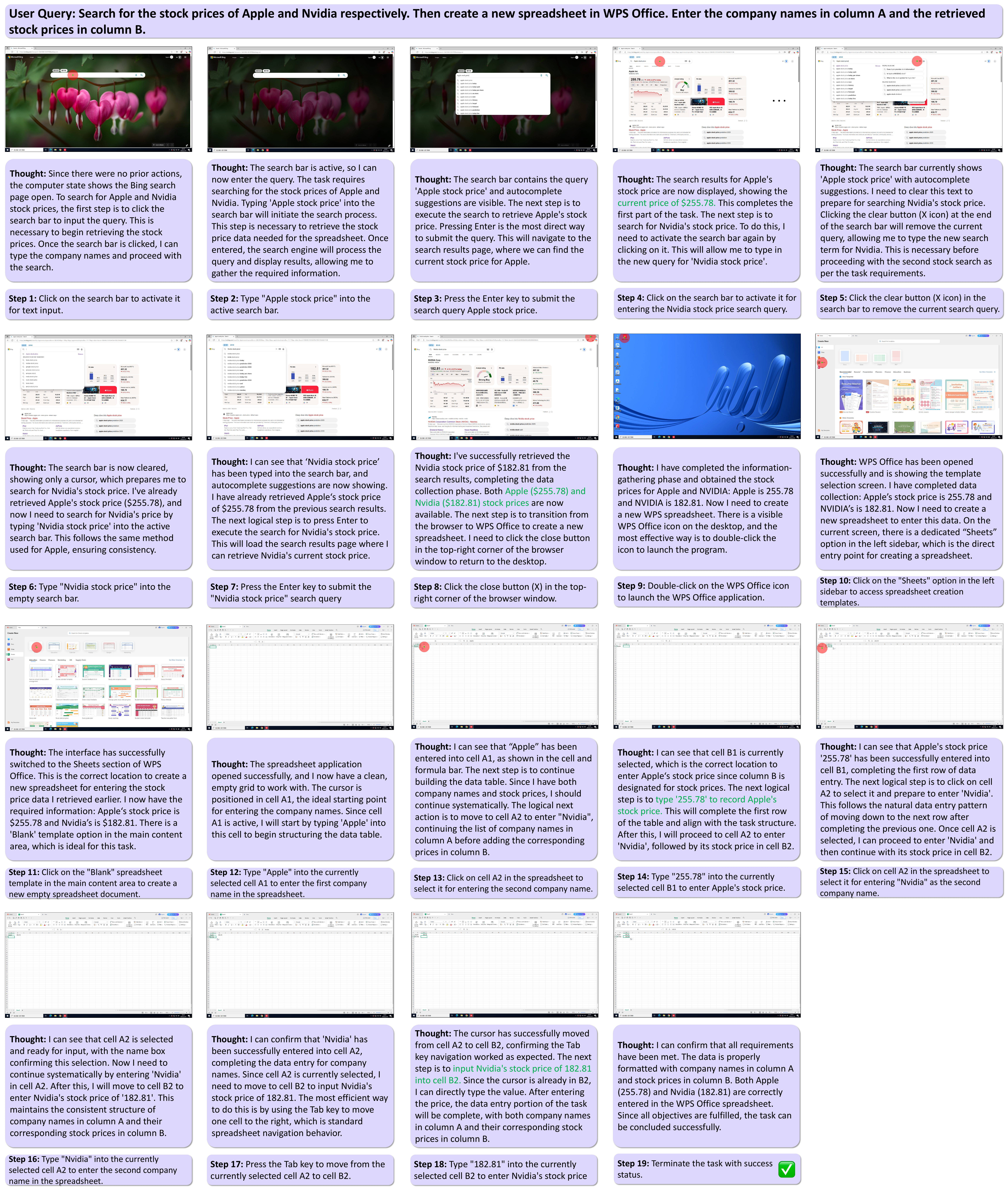}
    \vspace{-0.1cm}
    \caption{A complete operation process on the Windows platform, in which the user query requires the agent to memorize key on-screen information.}
    \label{fig:pc_memory}
    \vspace{-0.25cm}
\end{figure}

\textbf{Computer Use Case (\cref{fig:pc_memory}).} 
Figure~\ref{fig:pc_memory} illustrates a case of GUI-Owl-1.5 executing a \textit{web search and note-taking} task on the Windows platform. To fulfill the user query, the agent is required to accurately perform multiple web searches and extract key information relevant to subsequent steps from the search results, which is stored as memory within the thought content (highlighted in green). Subsequently, the agent switches to a different application, creates a new spreadsheet in WPS Office, and fills in the corresponding content at the appropriate cells based on the memorized information.
The thoughts generated by GUI-Owl-1.5 during the execution steps demonstrate its understanding of screen content, precise grounding, analysis of task progress, and memorization of key information, validating the effectiveness of our proposed unified CoT synthesis pipeline.
% In this case, the agent is instructed to ``\textit{Search for the stock prices of Apple and Nvidia respectively. Then create a new spreadsheet in WPS Office. Enter the company names in column A and the retrieved stock prices in column B.}'' The agent first opens a browser to search for Apple's stock price, memorizes the retrieved value, then searches for Nvidia's stock price and stores it as well. After collecting both pieces of information, the agent switches to WPS Office, creates a new spreadsheet, and accurately fills in the company names and their corresponding stock prices from memory. This case demonstrates the agent's ability to retain key information across application switches and long interaction horizons, and to precisely recall and apply memorized content in a structured downstream task—all without any external memory module.

\begin{figure}[!t]
    \centering
    \includegraphics[width=0.99\textwidth]{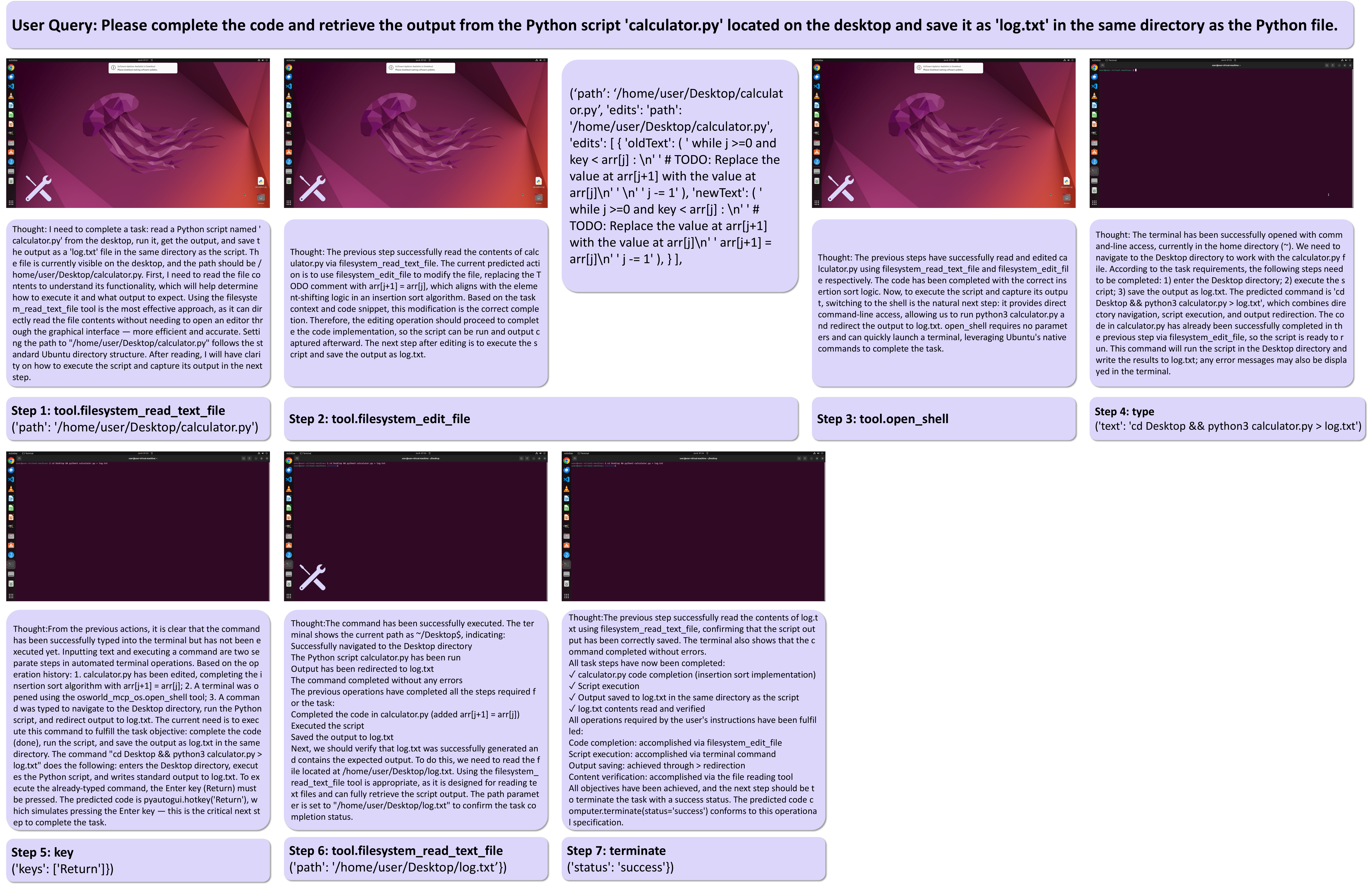}
    \vspace{-0.1cm}
    \caption{A case of a complete operation process on a desktop platform, which combining extended tools and computer use actions.}
    \label{fig:mcp_case}
    \vspace{-0.25cm}
\end{figure}

\textbf{Tool Use Case (\cref{fig:mcp_case}).} In this case, the agent is tasked with completing a partially implemented Python script on the desktop and saving its execution output. The agent seamlessly interleaves MCP tool calls with GUI operations: it first reads the source code via the \textit{filesystem\_read\_text\_file} tool, identifies and fixes the incomplete insertion sort implementation using \textit{filesystem\_edit\_file}, then opens a terminal through \textit{osworld\_mcp\_os.open\_shell} to execute the script via command-line input, and finally verifies the output by reading the generated log file. This case demonstrates GUI-Owl 1.5's ability to autonomously decide when to use tool invocation versus direct GUI manipulation within a single trajectory.

\section{Conclusion}
In this work, we presented GUI-Owl-1.5, the native GUI agent model that features instruct/thinking variants in multiple sizes (2B/4B/8B/32B/235B) and supports a range of devices (desktop, mobile, browser, and more). GUI-Owl-1.5 achieves state-of-the-art performance on 20+ GUI benchmarks, comprehensively covering GUI automation, grounding, tool calling, memory, and knowledge tasks. We innovatively improve the model’s robust generalization in real-world application scenarios through a Hybrid Data Flywheel, unified enhancement of agent capabilities, and multi-device environment RL scaling. We hope that the open-source release of GUI-Owl-1.5 will advance the adoption of GUI agents for device automation across a wide range of platforms.

\clearpage
\bibliography{iclr2024_conference}
\bibliographystyle{iclr2024_conference}

\clearpage
\appendix

\end{document}